\pdfoutput=1 % Force arXiv's AutoTeX to use pdflatex; the magic comment above is ignored there.
\documentclass{pretty-preprint}
\usepackage{xcolor}
\usepackage[section]{placeins} % Keep planned visuals inside their owning sections.
\usepackage{wrapfig}
\usepackage[numbers,sort&compress]{natbib}
\usepackage[table]{xcolor}

\definecolor{opensourcebg}{RGB}{230,245,234}
\definecolor{closedsourcebg}{RGB}{220,220,220}
\usepackage{xspace}

\newcommand{\eg}{\emph{e.g.,}\xspace}

\title{CausalWM: Causal Chain-of-Thought Reasoning for Embodied World Model}
\runningtitle{CausalWM}
\author[1,\textdagger]{Ziming Xu}
\author[1,2,\textdagger,$\ddagger$]{Shuang Liang}
\author[1]{Ruobing Han}
\author[1]{Ziqiao Xi}
\author[1,3,$\ddagger$]{Mingxing Rao}
\author[1,*]{Kun Zhou}
\author[1]{Zijun Zhang}
\author[1]{Yuchen Yan}
\author[1,2,$\ddagger$]{Yufan Wei}
\author[1]{Junbo Huang}
\author[1]{Yifei Shao}
\author[1,2,$\ddagger$]{Fang Nan}
\author[1]{Biwei Huang}
\affiliation[1]{Aether AI}
\affiliation[2]{University of California, San Diego}
\affiliation[3]{Vanderbilt University}
\contribution[*]{Corresponding author and project leader}
\contribution[\textdagger]{Equal contribution}
\contribution[$\ddagger$]{Work done during internship at Aether AI}

\brandmarkbox{\includegraphics[height=8.5pt]{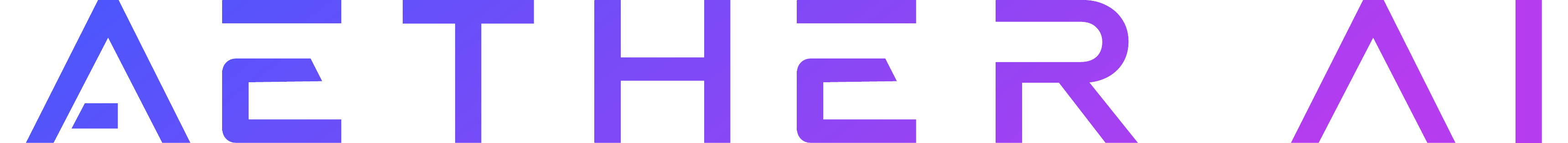}}

\correspondence{Kun Zhou (\email{franciskunzhou@gmail.com})}
\metadata[Code]{\href{https://github.com/AetherLabsAI/CausalWM}{github.com/AetherLabsAI/CausalWM}}
\metadata[Model Weights]{\href{https://huggingface.co/AetherLabs-AI/CausalWM}{huggingface.co/AetherLabs-AI/CausalWM}}
\metadata[Website]{\href{https://aetherlabsai.github.io/CausalWM/}{CausalWM}}

\abstract{
Embodied world models learn to predict future physical dynamics from visual observations and control signals, where physical knowledge is implicitly entangled within latent representations. %, making it difficult to determine whether key causal factors are faithfully captured. 
We introduce \textbf{CausalWM}, a 16B embodied world model that performs explicit \textbf{causal chain-of-thought reasoning} before future video prediction. CausalWM organizes useful variables into a reasoning trajectory, allowing the model to progressively capture causal dependencies underlying physical evolution. To train CausalWM, we collect 31K hours embodied data and develop a three-stage paradigm consisting of large-scale video pre-training, causal CoT mid-training, and multi-objective RL post-training. Despite using only a limited set of supervised CoT variables, CausalWM exhibits emergent in-context learning capabilities, enabling contextual visual feature guidance and efficient few-step generation. CausalWM achieves state-of-the-art performance across language-conditioned, action-conditioned, single-view and multi-view benchmarks, including Top-1 performance on TriWorldBench leaderboard.
}

\begin{document}
\maketitle

\begin{figure}[!htbp]
  \centering
  \includegraphics[width=\linewidth]{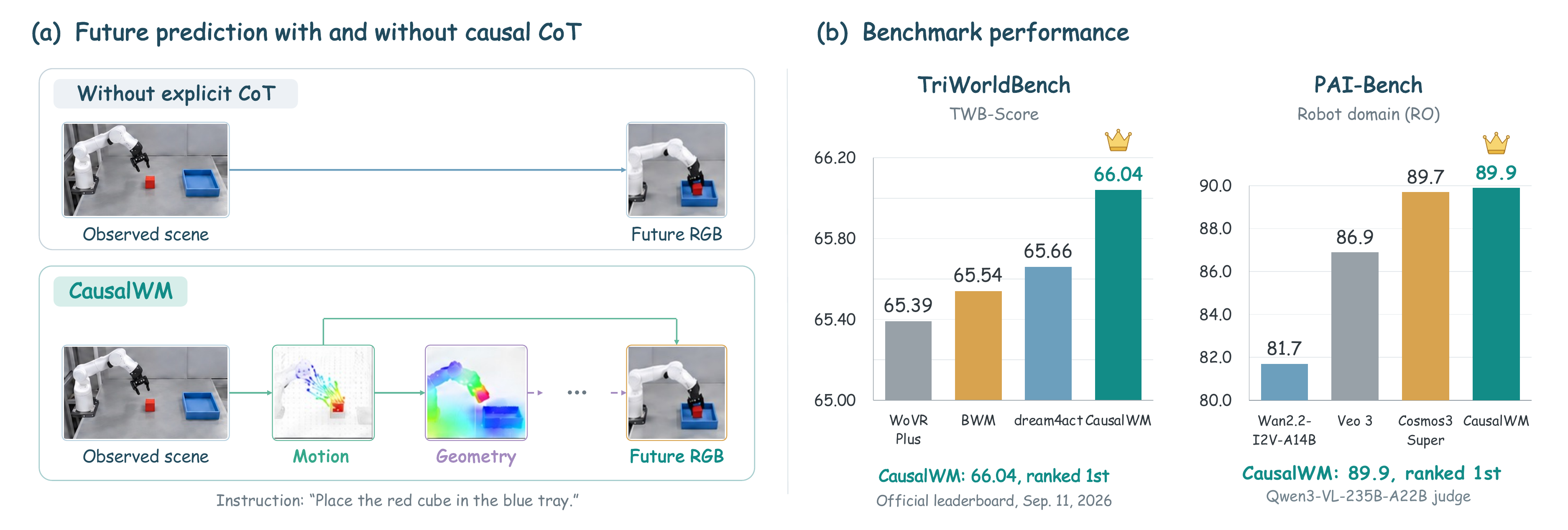}
  \caption{(a) Comparison of direct future video prediction and our CausalWM with an explicit chain of thought consisting of motion, geometry, and any useful representations. (b) CausalWM ranks 1st on action-conditioned multi-view TriWorldBench~\cite{triworldbench2026} as of Sep. 11, 2026, and reaches state-of-the-art performance on language-conditioned single-view PAI-Bench~\cite{paibench} under the Qwen3-VL judge.}
  \label{fig:rsi-overview}
\end{figure}

% Introduction
\section{Introduction}

Understanding and predicting how the physical world evolves is a fundamental capability for embodied intelligence. Recent embodied world models have shown promising progress in forecasting future visual observations conditioned on language instructions or low-level actions~\cite{yang2023learning,agarwal2025cosmos,abot-physworld,dreamdojo,qwenrobotworld,lingbot-video,cosmos3}. 
By pre-training on large-scale video data, embodied world models can capture rich visual dynamics and provide a predictive interface for downstream applications such as action simulation, robot control and planning~\cite{dreamzero,lingbot-va,vera,cosmos-policy,dreamdojo,ctrl-world}. 
Typically, existing embodied world models aim to learn a state transition function that maps the current visual context and control signal to future observations, allowing the model to implicitly acquire physical knowledge and action-conditioned dynamics from data~\cite{qwenrobotworld,dreamdojo,abot-physworld,lingbot-video,cosmos3,irasim,vid2world,ivideogpt}. 

However, such physical knowledge is often entangled within latent model representations during future video prediction. Thus, it is unclear whether key causal variables have been faithfully captured by the model or simply bypassed through shortcut correlations~\cite{kang2024far,cdlam,yocausal,motamed2026generative,geirhos2020shortcut,scholkopf2021toward}. 
This issue becomes more pronounced as the prediction horizon grows or the scene becomes more complex~\cite{self-forcing,xue2026acwm}.
In contrast, real-world physical evolution is inherently structured by a sequence of causally dependent changes (\eg objects move before contact, contacts induce motion change)~\cite{yi2019clevrer,li2020causal,baradel2019cophy,chen2022comphy,wang2026chain,dit-mem}. 
These variables naturally form a causal graph to capture not only what future state may occur, but also how that state emerges from the underlying physical dynamics.

Learning such causal structures is crucial for embodied world models, as they provide a foundation for understanding the physical world. A key challenge is to capture the causal dependencies among useful variables involved in physical dynamics. Chain-of-thought (CoT) reasoning offers a natural perspective for addressing this problem~\cite{wang2026chain,tang2026causal}. 
In language models, CoT improves complex problem solving by decomposing a difficult prediction into a sequence of intermediate reasoning steps~\cite{wei2022-llm-cot,zhou2022least}, 
then progressively derives the final prediction. We argue that CoT is particularly suitable for embodied world modeling, where physical environments contain a rich set of useful variables, \eg optical flow, object trajectories~\cite{flowwam,zhuang2026causalmotion,gao2025flip}. 
These variables explicitly characterize complementary aspects of physical evolution that are hidden in raw visual observations, providing natural building blocks for modeling causal relationships in physical simulation.

Motivated by this observation, we introduce \textbf{CausalWM}, a 16B embodied world model that performs \textbf{causal chain-of-thought reasoning} before predicting future visual observations. 
Concretely, CausalWM organizes useful physical variables into an explicit reasoning trajectory that describes how the current physical state evolves toward the future. 
Each step in the causal CoT captures a meaningful intermediate transition, enabling the model to progressively reason from the current observation and control signal to the resulting future state. 
In this way, future prediction becomes a structured process in which intermediate physical changes connect the initial cause to the final visual consequence. 
Moreover, progressively predicting these intermediate variables provides auxiliary supervision that encourages the model to better capture the intrinsic causal structure underlying physical dynamics.

To build CausalWM, we collect 31K hours of embodied data from diverse sources and develop a three-stage training paradigm. 
We first conduct large-scale pre-training to learn pixel-level generation of embodied videos and establish a strong visual dynamics prior. We then perform mid-training with causal CoT reasoning, where we select a small set of variables and organize them from simple to complex into an explicit reasoning trajectory. 
Finally, we apply post-training with multi-objective reinforcement learning, using verifiable rewards computed from the final predicted videos to further optimize physical consistency and generation quality.

Notably, although the CoT is constructed from only a limited set of variables, the resulting reasoning capability enables CausalWM to causally attend to a broader range of in-context features. 
This emergent in-context learning capability further gives rise to new model behaviors, including in-context visual feature guidance and fast video generation with extremely few denoising steps. 
Benefiting from these capabilities, CausalWM achieves Top-1 performance on TriWorldBench and reaches state-of-the-art performance on the robot domain of PAI-Bench.

We summarize our contributions as follows:
\begin{itemize}
\item We introduce CausalWM, a new embodied world model that formulates future prediction as causal chain-of-thought reasoning, explicitly organizing useful variables into intermediate reasoning steps to better capture the causal structure.

\item We develop a three-stage training paradigm built on 20K hours of diverse embodied data, consisting of large-scale pre-training, causal CoT mid-training, and multi-objective RLVR for post-training.

\item CausalWM exhibits strong emergent in-context reasoning capabilities beyond the explicitly supervised CoT variables, enabling in-context visual feature guidance and efficient few-step video generation.

\item CausalWM achieves state-of-the-art performance on language-conditioned, action-conditioned, single-view and multi-view embodied world model benchmarks, including Top-1 results on TriWorldBench leaderboard.
\end{itemize}

% Preliminary
\section{Preliminary}
\label{sec:preliminary}

We formulate embodied world model and chain-of-thought reasoning for future prediction.

\paragraph{Embodied World Model.}
\label{sec:prel-embodied-wm}
An embodied world model aims to simulate how the physical world evolves under external controls, providing a predictive model of future visual dynamics for embodied planning, control, and decision making~\cite{hafner2019learning}. Given the current visual observation, or a history of past observations, the model predicts how the scene will evolve when following a specified control signal. To support different downstream requirements, we consider two common forms of control: \emph{language control}, which specifies a desired behavior or semantic instruction, and \emph{action control}, which directly specifies low-level physical actions~\cite{yang2023learning}. Concretely, let $\boldsymbol{o}_{1:t}$ denote the observed visual history up to time $t$, and $\boldsymbol{c}$ denote the corresponding control signal, either a language instruction $\boldsymbol{l}$ or an action sequence $\boldsymbol{a}_{t:t+H-1}$. An embodied world model parameterized by $\theta$ aims to predict the future visual dynamics over a horizon of $H$ steps:
\begin{equation}
p_\theta\!\left(
\boldsymbol{o}_{t+1:t+H}
\mid
\boldsymbol{o}_{1:t}, \boldsymbol{c}
\right),
\qquad
\boldsymbol{c}\in\{\boldsymbol{l},\,\boldsymbol{a}_{t:t+H-1}\}.    
\end{equation}

\paragraph{Chain-of-thought Reasoning.}
\label{sec:prel-cot}
Chain-of-thought (CoT) reasoning decomposes a complex prediction problem into a sequence of intermediate reasoning steps before producing the final output. By explicitly modeling these intermediate variables, CoT provides additional structure for capturing multi-step dependencies. Formally, given an input $\boldsymbol{x}$ and target output $\boldsymbol{y}$, CoT introduces an intermediate reasoning sequence $\boldsymbol{r}=(r_1,r_2,\ldots,r_K)$ and factorizes the joint prediction as

$$
p_\theta(\boldsymbol{y},\boldsymbol{r}\mid\boldsymbol{x})
=
\left[
\prod_{k=1}^{K}
p_\theta(r_k\mid\boldsymbol{x},r_{<k})
\right]
p_\theta(\boldsymbol{y}\mid\boldsymbol{x},\boldsymbol{r}),
$$

This formulation allows the final prediction to be progressively derived through a structured reasoning trajectory. In embodied world modeling, various intermediate physical variables, such as optical flow and depth, have been widely exploited to improve the modeling of scene dynamics~\cite{gao2025flip,zhen2025tesseract}. These variables capture complementary aspects of physical change that are often implicit in raw visual observations. In this work, we incorporate such informative variables as explicit chain-of-thought reasoning steps, enabling the world model to reason through physically meaningful intermediate transitions before generating the future visual state.

% Data section
\section{Data} % Ruobing
\label{sec:data}

\subsection{Data Collection} 
\label{sec:data-collection}
%\todo{Add the original dataset/report citations for every dataset named in this subsection and Table~\ref{tab:data-hours}.}

We assemble a broad collection of interaction/manipulation videos spanning human egocentric activity, real-robot demonstrations, and simulated manipulation. Our current inventory covers 20 source families and approximately 31,000 hours before preprocessing. Table~\ref{tab:data-hours} summarizes the details. Selected subsets or constituent datasets are used for some sources.
%Three complementary types of sources below expose the world model to both the shared structure of object interaction, and the visual and kinematic variation associated with different embodiments.

\paragraph{Human Egocentric Videos.}
Egocentric-10K~\cite{data_egocentric10k} contributes 192,504 videos totaling 9,980.3 hours, providing the largest individual source in the inventory before filtering. Ego4D~\cite{data_ego4d}, EgoDex~\cite{data_egodex}, EPIC-Kitchens~\cite{data_epic_kitchens}, and H2O~\cite{data_h2o} complement this scale with daily activities, dexterous manipulation, tool use, and hand-object interaction. EgoVerse~\cite{data_egoverse} provides both human egocentric videos, and also robot demonstrations of similar bimanual manipulations.

\paragraph{Real Robot Videos.}
AgiBot-World Beta~\cite{data_agibot_beta} provides substantial robot interaction coverage, complemented by AgiBot-World 2026~\cite{data_agibot_2026}, Galaxea~\cite{data_galaxea}, RoboCOIN~\cite{data_robocoin}, RoboMIND~\cite{data_robomind}, DROID~\cite{data_droid}, RT-1~\cite{data_rt1}, BridgeData V2~\cite{data_bridgev2}, selected constituent OXE datasets~\cite{data_oxe}, Humanoid-Everyday~\cite{data_huamnoid_everyday}, and RoVid-X~\cite{data_rovidx}. These sources span single-arm, bimanual, mobile-manipulation, and humanoid platforms, with varied end effectors and camera placements, and contain demonstrations with varying arm geometry, workspace layout, and execution patterns.

\paragraph{Simulation Data.}
Simulation data complements real-world data by supplying combinations of embodiment, objects, trajectories and scenes that are costly or hard to collect, while retaining explicit embodiment structure, textual and numerical annotations. InternData-A1~\cite{data_interna1}, RoboCasa365~\cite{data_robocasa365}, RoboTwin 2.0~\cite{data_robotwin2}, and the digital-twin data accompanying AgiBot-World 2026~\cite{data_agibot_2026} cover diverse tasks, object arrangements, scenes, and embodiments. InternData-A1 includes Franka, Lift-2, Split ALOHA, and Genie-1. RoboTwin 2.0 further covers ALOHA AgileX, ARX X5, Franka, and UR5.

\begin{table}[!t]
  \centering
  \begingroup
  \fontsize{9}{11}\selectfont
  \setlength{\tabcolsep}{3pt}
  \renewcommand{\arraystretch}{1.13}
  \newcommand{\dataembodimentrule}{\noalign{\vskip\aboverulesep\hbox to\linewidth{\normalfont\fontsize{6}{6}\selectfont\dotfill}\vskip\belowrulesep}}
  \newcommand{\dataembodimentgroup}[2]{%
    \begingroup
    \setbox0=\hbox{\strut#2}\dimen6=\wd0 \dimen0=\arraystretch\baselineskip
    \dimen2=\dimexpr\dimen0*#1/2-\dimen0/2\relax
    \dimen4=\dimexpr\dimen0*#1/2-10pt\relax
    \makebox[\linewidth][l]{\hspace{8pt}\raisebox{\dimexpr2pt-\dimen2\relax}[0pt][0pt]{%
      \raisebox{-.5\height}{\vbox{\offinterlineskip
        \hbox to\dimen6{\hfil\rule{0pt}{\dimen4}\hfil}\kern3pt
        \hbox to\dimen6{\hfil\strut#2\hfil}\kern3pt
        \hbox to\dimen6{\hfil\rule{0pt}{\dimen4}\hfil}}}}}%
    \endgroup}
  \begin{tabularx}{\linewidth}{@{}>{\raggedright\arraybackslash}p{0.255\linewidth}
    >{\raggedleft\arraybackslash}p{0.140\linewidth}
    >{\raggedleft\arraybackslash}p{0.140\linewidth}
    >{\raggedright\arraybackslash}X
    >{\raggedright\arraybackslash}p{0.130\linewidth}@{}}
    \toprule[1pt]
    \textbf{Dataset} & \textbf{Source (h)} & \textbf{Retained (h)} & \hspace{8pt}\textbf{Embodiment} & \textbf{Views} \\
    \midrule
    Ego4D~\cite{data_ego4d} & 3,795.9 & 798.5 & \dataembodimentgroup{5}{Human hands} & Ego \\
    Egocentric-10K~\cite{data_egocentric10k} & 9,980.3 & 6,878.2 &  & Ego \\
    EgoDex~\cite{data_egodex} & 829.4 & 513.4 &  & Ego \\
    EPIC-Kitchens~\cite{data_epic_kitchens} & 69.4 & 16.4 &  & Ego \\
    H2O~\cite{data_h2o} & 1.1 & 0.4 &  & Ego + ext. \\
    \dataembodimentrule
    BridgeData V2~\cite{data_bridgev2} & 87.0 & 59.0 & \dataembodimentgroup{5}{Single-arm} & Ext. + wrist \\
    DROID~\cite{data_droid} & 273.1 & 189.4 &  & Ext. + wrist \\
    OXE~\cite{data_oxe} & 373.6 & 336.5 &  & Mixed \\
    RoboCasa365~\cite{data_robocasa365} & 1,666.3 & 1,502.9 &  & Ext. + wrist \\
    RT-1~\cite{data_rt1} & 350.6 & 247.5 &  & Ego \\
    \dataembodimentrule
    AgiBot-World 2026~\cite{data_agibot_2026} & 284.2 & 220.9 & \dataembodimentgroup{6}{Dual-arm} & Ego + wrist \\
    AgiBot-World Beta~\cite{data_agibot_beta} & 2,245.8 & 2,022.1 &  & Ego + wrist \\
    Galaxea~\cite{data_galaxea} & 540.9 & 485.9 &  & Ego + wrist \\
    Humanoid-Everyday~\cite{data_huamnoid_everyday} & 31.8 & 28.5 &  & Ego \\
    RoboCOIN~\cite{data_robocoin} & 489.0 & 488.0 &  & Mixed \\
    RoboTwin 2.0~\cite{data_robotwin2} & 162.5 & 113.8 &  & Ext. + wrist \\
    \dataembodimentrule
    InternData-A1~\cite{data_interna1} & 3,071.1 & 1,507.6 & \dataembodimentgroup{3}{Single- and dual-arm} & Mixed \\
    RoboMIND~\cite{data_robomind} & 329.3 & 150.5 &  & Mixed \\
    RoVid-X~\cite{data_rovidx} & 5,473 & 4,004.6 &  & Mixed \\
    \dataembodimentrule
    EgoVerse~\cite{data_egoverse} & 1,176.6 & 352.2 & \makebox[\linewidth][l]{\hspace{8pt}Human hands + robot arms} & Mixed \\
    \midrule
    \textbf{Total} & \textbf{31,230.8} & \textbf{19,916.2} & \multicolumn{2}{l}{20 sources} \\
    \bottomrule[1pt]
  \end{tabularx}
  \endgroup
  \caption{Source hours, retained hours, and domains of our collected video data pool. Hours refer to the collected input pool and prepared clips, respectively, with synchronized views of the same trajectory counted only once. Selected subsets or constituent datasets are used for some sources. \\ DROID: the DreamZero-DROID release~\cite{dreamzero}.}% Source hours report local inventories. Retained hours sum prepared clips or steps, with synchronized views counted once. %\textsuperscript{\dag} OXE and RoVid-X source hours refer to our selected subset. 
  %Views describe available camera placements.}
  \label{tab:data-hours}
\end{table}

\begin{figure}[!t]
  \centering
  % Vector PDF exported from the current editable PPT; SVG source is also in assets.
  \includegraphics[width=\linewidth]{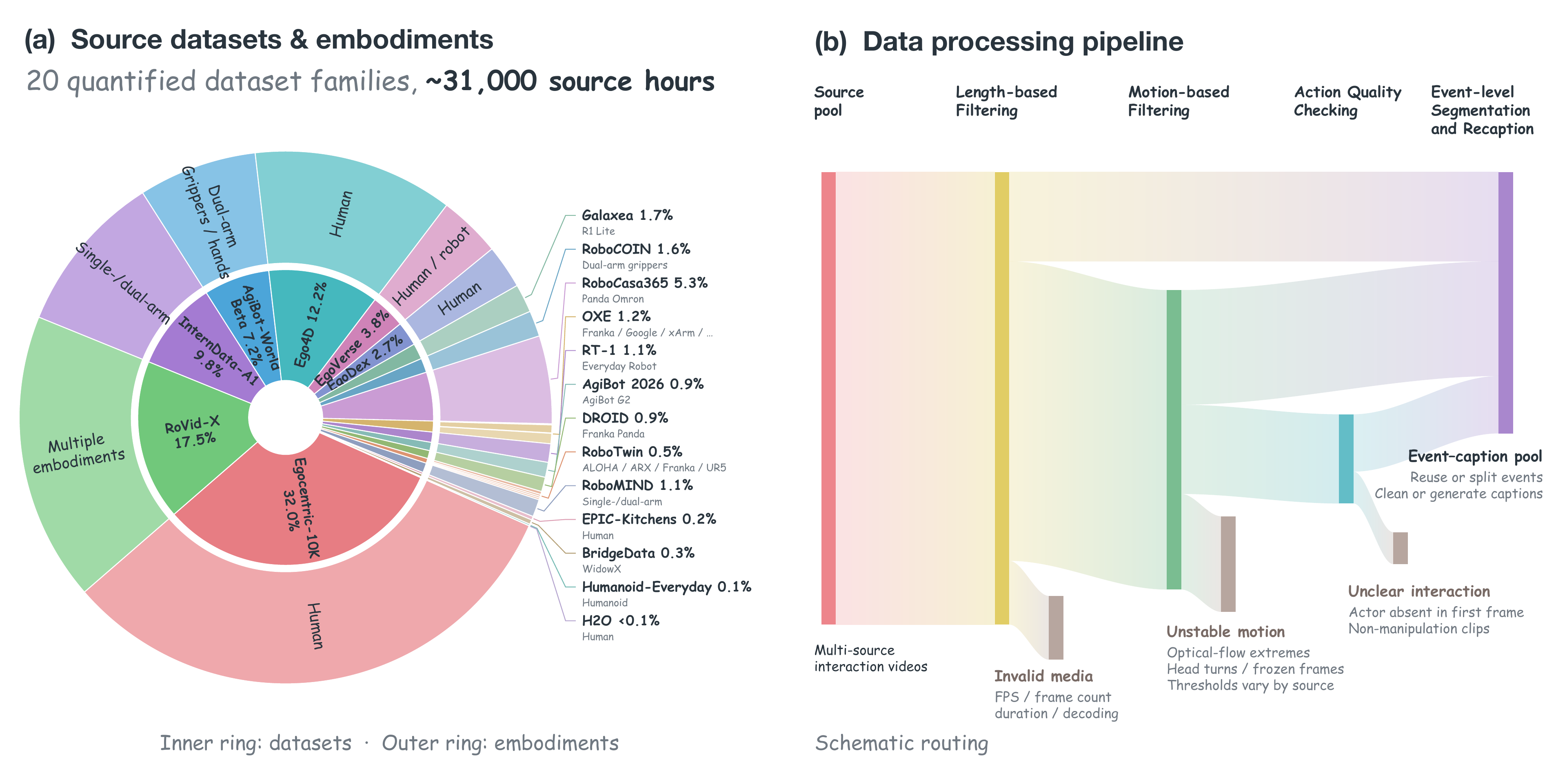}
  \caption{\textbf{Data collection and preprocessing pipeline.}
  (a) Composition of the current source pool, with dataset families in the inner ring and their embodiments in the outer ring. Details are given in Table~\ref{tab:data-hours}.
  (b) Preprocessing pipeline for filtering, checking and event-level reconstruction. Flow widths are schematic.}
  \label{fig:data-curation}
\end{figure}

\subsection{Data Processing}
\label{sec:data-processing}
For data processing, we first filter out low-quality samples based on abnormal duration, undesirable motion patterns, and poor action quality, and then perform temporal segmentation and re-captioning to standardize instruction granularity. Figure~\ref{fig:data-curation} summarizes the composition of our data mixture and the overall data processing pipeline.

%\textit{Quality Filtering.}
%Human inspection of representative videos and annotations assesses visual clarity, camera stability, interaction density, and the availability and reliability of native steps and captions. This inspection determines the processing route for each dataset or its subset. High-quality videos with clean events and captions, such as those from AgiBot-World Beta, are shortlisted for direct reuse after format normalization. Other recordings undergo source-specific filtering, with the steps that we describe below.

\paragraph{Length-based Filtering.}
We filter candidate clips according to their temporal validity and usable duration. Frame rate, frame count, timestamps, and interval length are jointly checked to reject corrupt media, invalid segments, clips with insufficient frames, or intervals that are too short for the required temporal sampling. Overly long recordings are either discarded or split into shorter candidates depending on the source. These constraints are adapted to the native frame rate of each dataset so that valid low-frame-rate robot demonstrations are preserved.
%Frame rate, frame count, duration, and interval bounds determine whether a candidate supports the required temporal sampling. Corrupt media, invalid timestamps, insufficient frames, and intervals that are too short are rejected; overly long recordings are excluded or split into manageable candidates depending on source-specific length requirements. Frame-count requirements are interpreted together with the source frame rate, preserving usable low-frame-rate robot demonstrations. These checks ensure that subsequent motion and semantic decisions operate on temporally valid and observable video.

\paragraph{Motion-based Filtering.}
We use optical-flow statistics to filter clips with undesirable motion patterns. Excessive flow often indicates head turns, camera shake, or large ego-motion, while extremely low motion can correspond to frozen or highly repetitive frames. We therefore apply source-specific motion thresholds according to camera characteristics and recording quality. Noisy egocentric datasets such as Egocentric-10K receive stronger filtering, while cleaner demonstrations such as EgoDex use lighter thresholds and stable-camera robot datasets may bypass this stage. For Egocentric-10K, this procedure removes the highest-flow 30\% of 8,952,263 candidate clips, retaining 6,266,584 intervals that yield approximately 6,878.2 hours of prepared clips.
%Optical-flow magnitude summarizes image motion across sampled frames. Source-specific thresholds suppress extreme motion associated with head turns, camera shake, or large camera-base displacement and help screen for frozen video or excessive repeated frames. These clips are typical in egocentric datasets, and can dominate otherwise useful interaction footage. Noisy sources such as Egocentric-10K receive stronger filtering; clearer egocentric demonstrations such as EgoDex receive much lighter selection, and stable-camera robot recordings can bypass this motion gate. For Egocentric-10K, optical-flow filtering removed the highest-flow 30\% of 8,952,263 candidate cuts, retaining 6,266,584 intervals totaling approximately 6,975.4 interval hours before packing. Motion filtering reduces visual artifacts that could confound the learning of true physical dynamics.

\paragraph{Action Quality Checking.}
We verify whether each clip contains a valid and task-relevant physical interaction. The initial frames are checked for recognizable hands, arms, grippers, or other relevant end effectors, and clips without an identifiable actor are removed. We further inspect task labels and instructions to retain manipulation-related activities while excluding clips dominated by non-interactive or irrelevant behaviors.
%The initial frame is checked for a recognizable hand, arm, gripper, or other relevant end effector. Clips without an identifiable actor are rejected, while recognizable partial views are retained. Task labels and instructions are also screened for relevance to object manipulation. Clips associated with non-manipulation activities are excluded.

\begin{figure}[!htbp]
  \centering
  \includegraphics[width=\linewidth]{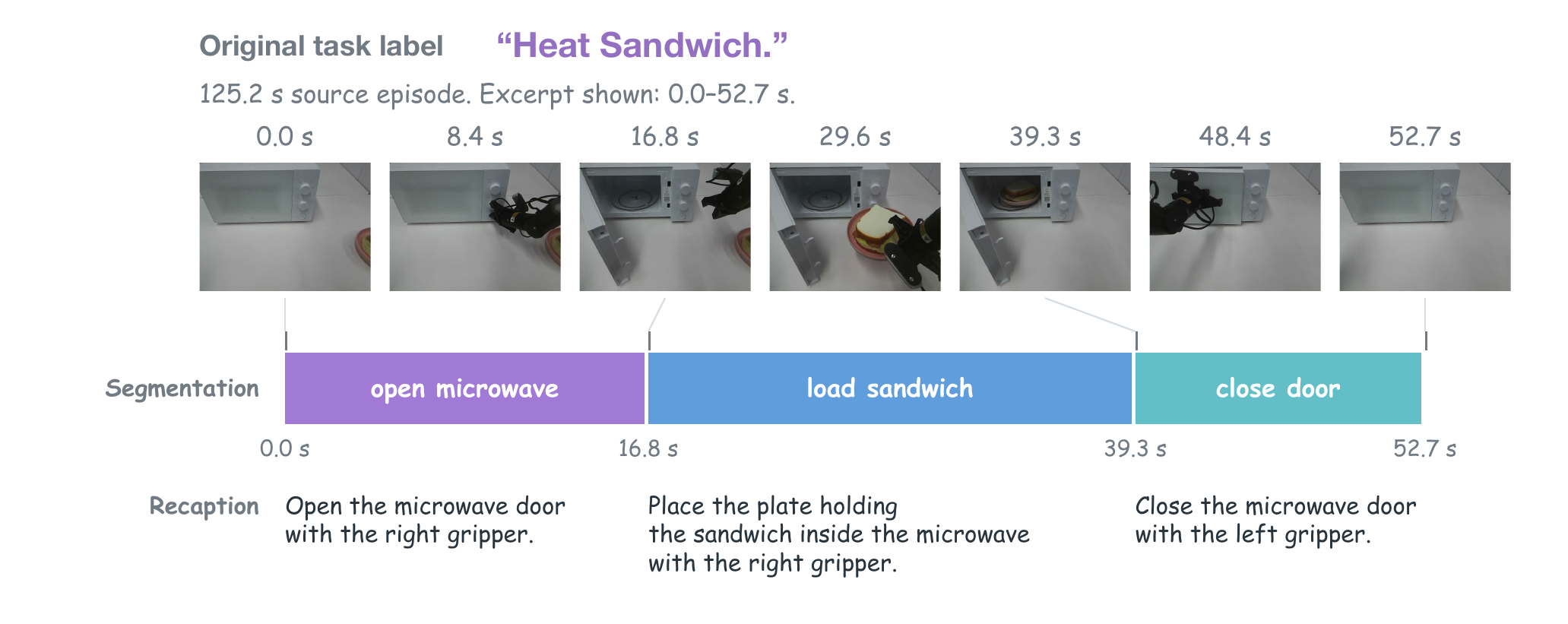}
  \caption{\textbf{Event-level video segmentation and caption cleaning.} An example from RoboCOIN~\cite{data_robocoin} with the raw instruction \emph{``Heat Sandwich.''} is segmented into three contiguous events, and assigned with new captions.} %Seven frames show the event boundaries and intermediate manipulation states; the timeline marks the three intervals.}
  \label{fig:data-cap-clean}
\end{figure}

\paragraph{Event-level Segmentation and Recaption.}
To align the temporal scope of each video clip with the granularity of its instruction, we further segment accepted videos into contiguous event-level clips and standardize their captions. This reduces ambiguity from long episodes containing multiple actions and provides cleaner supervision for learning action-conditioned dynamics. When reliable native step annotations or captions are available, we reuse them after normalization into concise, verb-led imperative descriptions without unnecessary environmental details. For sources without suitable annotations, we segment videos using available temporal labels and interaction cues, with VLM assistance when necessary, and generate a cleaned caption for each resulting interval. As illustrated in Figure~\ref{fig:data-cap-clean}, each caption focuses on a single event, specifying the manipulated object and its destination when applicable while maintaining consistent references to the embodiment and objects. LLMs are used to remove placeholder tokens and correct obvious wording errors without changing the underlying action semantics, while VLMs generate concise action descriptions for sources lacking usable captions.

%In the illustrated example, the first 52.7 seconds of a 125.2-second episode are segmented into three events:
%\begin{center}
%\textbf{\emph{Open the microwave door with the gripper.}}\quad (0.0--16.8 s)[0.5em]
%\textbf{\emph{Place the plate holding the sandwich inside the microwave.}}\quad (16.8--39.3 s)[0.5em]
%\textbf{\emph{Close the microwave door with the gripper.}}\quad (39.3--52.7 s)
%\end{center}
%The resulting captions preserve a concise and consistent imperative format, focusing on the event action, manipulated object, and destination when applicable.

% This example was manually prepared from RoboCOIN episode 000000 (Agilex_Cobot_Magic_heat_sandwich); not representative of corpus-wide caption normalization.
 
% kept as separate file for easier management
% ----- Moved to the file ----- %
% \section{Data} % Ruobing
% \label{sec:data}
% \subsection{Data Collection}
% proportion figure, raw data sources
% \subsection{Data Processing}
% Event-level, Filtering, Recaption
% ----------------------------- %

% Model
\begin{figure}[!htbp]
  \centering
  \includegraphics[width=\linewidth]{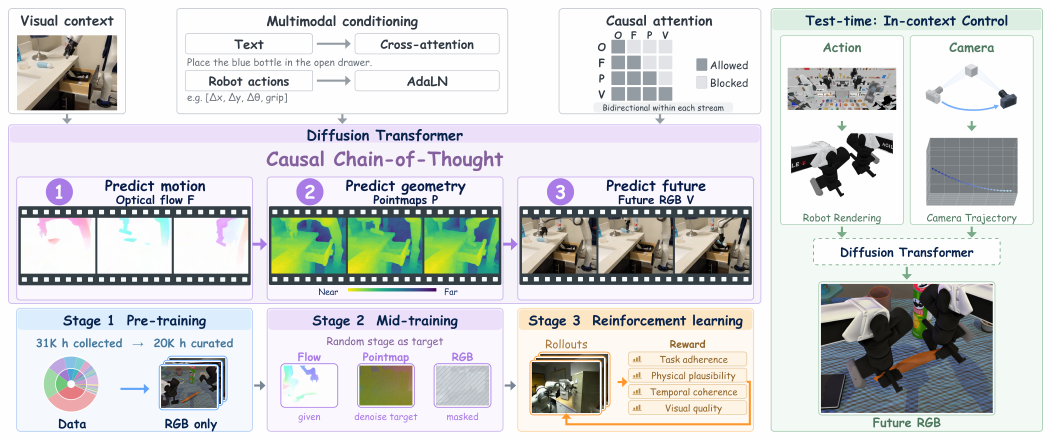}
  \caption{\textbf{Overview of CausalWM.} It supports either language or robot action conditions, and can performs causal chain-of-thought reasoning by sequentially predicting the optical flow, pointmaps, and finally the future RGB video. A causal attention mask is to ensure the undirectional conditional information flow, and all tasks reuse the diffusion Transformer. After learning the causal CoT ability, CausalWM can efficiently learn flexible control guidance through in-context features.}
  %\textbf{Top:} Visual context and multimodal conditions enter a shared diffusion Transformer. Observed RGB frames, history, or aligned visual features are encoded by a frozen VAE into fixed context tokens. Text conditions enter each Transformer block through cross-attention. Robot actions follow two pathways: an action bridge injects latent-action chunks through AdaLN, and the same actions are rendered into control videos via URDF forward kinematics, encoded by the same frozen VAE, and fused with the RGB latents through channel concatenation and a learned projection, adding no extra attention tokens. Further modalities can enter through the same interface. \textbf{Bottom:} In the illustrated language-conditioned rollout, the same Transformer successively denoises optical flow $F$, pointmaps $P$, and future RGB $V$; each completed stream is reused as context for later stages. Causal attention enforces the order $O\!\rightarrow\!F\!\rightarrow\!P\!\rightarrow\!V$ while remaining bidirectional within each stream. RGB, flow, pointmaps, and control videos share one frozen VAE. Visualizations are illustrative.}
  \label{fig:cwm-overview}
\end{figure}

\section{Model} % Ziming & Shuang
\label{sec:method}

We first introduce the model architecture of our CausalWM, and then multi-stage training for learning to perform causal chain-of-thought reasoning for embodied world modeling.
Figure~\ref{fig:cwm-overview} illustrates the overviews of our CausalWM.
%causal chain-of-thought in CausalWM: the model predicts motion and geometry before generating future RGB, with the observation and instruction conditioning every step.

\subsection{Model Architecture}
\label{sec:model-architecture}
Our CausalWM is built on the diffusion Transformer architecture that supports learning in-context conditions. During inference, it can perform stage-wise denoising for chain-of-thought reasoning.

\subsubsection{Unified Diffusion Transformer for In-context Conditioning}
Our world model is built on a Diffusion Transformer (DiT) backbone designed to support diverse control signals through a unified in-context conditioning interface. 
Given an observed video clip $\boldsymbol{o}_{1:T}$, a spatio-temporal VAE encoder first maps the visual observations into latent frames $\boldsymbol{z}_{1:T}$. Starting from noisy future latents, the DiT performs iterative denoising to generate the future latent frames, which can be decoded to pixels by the VAE decoder.  Following common practice in flow-matching models~\cite{esser2024scaling, lee2024improving, rao2026generalization}, we bias the training timestep distribution toward the high-noise regime: $t$ is drawn from a logit-normal distribution $t=\sigma(\mu+\epsilon),\ \epsilon\sim\mathcal{N}(0,1)$, where the shift $\mu$ grows linearly with the number of latent tokens, and the samples are rescaled to cover $[0,1]$; with probability 0.1 we instead sample $t$ uniformly to retain coverage of the low-noise end. Similar to existing world models~\cite{yang2023learning,abot-physworld,lingbot-video}, our DiT supports both language and action conditioning, through cross-attention and adaptive layer normalization (AdaLN), respectively~\cite{rombach2022high,dit,lingbot-world,irasim}. 
%Language features are injected through cross-attention, where visual tokens attend to the encoded instruction representations~\cite{}, while action controls are incorporated through adaptive layer normalization (AdaLN), which modulates the hidden activations with action-dependent scale and shift parameters~\cite{}.

Beyond these explicit control interfaces, our backbone further supports \emph{in-context conditioning}. Concretely, useful auxiliary features can be encoded as the observed video, then concatenated with the video tokens to form a unified context sequence. These context tokens are jointly processed by the DiT through self-attention. In this way, the model can directly use auxiliary information when predicting future observations:

\begin{equation}
\mathcal{L}_{\mathrm{diff}}
=
\mathbb{E}_{\sigma,\boldsymbol{\epsilon}}
\left[
\left\|
\boldsymbol{v}_{\theta}
\!\left(
\boldsymbol{z}_{t+1:t+H}^{\sigma}
\,\middle|\,
\boldsymbol{z}_{1:t},
\hat{\boldsymbol{c}},
\boldsymbol{c},
\sigma
\right)
-
\boldsymbol{v}_{\sigma}
\right\|_2^2
\right],
\end{equation}

where $\hat{\boldsymbol{c}}$ denotes the auxiliary control through in-context conditioning, $\boldsymbol{c}$ denotes the explicit language or action condition and $\boldsymbol{v}_{\sigma}$ is the target denoising velocity at noise level $\sigma$. This formulation makes the conditioning interface highly flexible: useful features can be introduced as additional context without modifying the backbone architecture. %Such a unified way provides the foundation for our causal chain-of-thought reasoning.

\subsubsection{Variable-by-Variable Denoising for Causal Chain-of-thought Reasoning}
Building on the unified in-context conditioning mechanism above, we further perform causal chain-of-thought reasoning by progressively predicting intermediate variables before generating the future video. Concretely, we organize multiple useful variables into an ordered reasoning sequence, where each variable is generated from the preceding context and subsequently reused for later prediction. In this way, future generation is decomposed into a sequence of physically meaningful transitions to gradually map the observation to future RGB frames.
To enforce this causal dependency, we introduce a causal attention mask that blocks shortcut paths from later streams to earlier predictions. Attention remains bidirectional within each stream, while the causal order is strictly enforced across streams. This prevents information leaking and encourages the model to follow the intended reasoning trajectory.
%Concretely, we use optical flow and pointmaps as intermediate variables, ordered from motion to geometry before future RGB generation. Let $\mathcal{C}=(E(\mathbf{x}_t),\mathbf{c})$ denote the encoded observation and control signal, and let $\mathbf{z}^{F}$, $\mathbf{z}^{P}$, and $\mathbf{z}^{V}$ denote the latent representations of optical flow, pointmaps, and future RGB video, respectively. The causal CoT is factorized as

At inference, we perform variable-by-variable denoising using the same causal order. We first denoise the first variable conditioned on the observed video and language or action control signal. The completed first variables are then fixed and inserted back into the context to guide next variable generation. After completing the chain of thought, all are reused as in-context conditions for denoising the future RGB video. For example, we can first predict the optical flow for motion feature modeling, then pointmap for geometry feature modeling, and finally the RGB video. Such a variable-by-variable generation process allows intermediate predictions to progressively constrain subsequent dynamics and guide the model toward more physically consistent future generation.

%\begin{figure}[htbp]
%\centering
%\includegraphics[width=.38\linewidth]{assets/cwm_causal_mask.pdf}
%\caption{\textbf{Stage-ordered attention.} Rows denote queries and columns denote keys/values. $O$ denotes observed RGB tokens; $F_i$, $P_i$ and $V_i$ denote spatial-token groups for flow, pointmaps and future RGB at latent-time index $i$, with $h$ denoting the final index. Filled cells permit attention; white cells are masked. Colors identify query groups. Intermediate time indices and the fixed initial flow group are omitted for clarity. Attention is bidirectional within each generated stream and follows the order $F\rightarrow P\rightarrow V$ across streams. Training and inference use the same visibility pattern.}
%\label{fig:cwm-mask}
%\end{figure}

\subsection{Multi-stage Training}
\label{sec:multi-stage-training}
To efficiently and effectively learn the causal CoT reasoning ability, we first perform large-scale pre-training for learning pixel-level embodied video generation, then mid-training for learning to perform causal chain-of-thought reasoning, finally post-training using multi-objective RL.

\subsubsection{Stage 1: Pre-training for Pixel-level Video Generation}
The first stage establishes the basic video-generation capability of CausalWM by training it to predict future observations in pixel space under both language and action conditions.

\textit{Language-conditioned video generation.} Given a language instruction, we encode and inject it into each DiT block through cross-attention. The model is optimized with a standard flow-matching objective over the generated frames. To support both single-frame prediction and history-conditioned continuation, each training clip is randomly divided into context and future frames, with independently sampled noise levels. The history is occasionally kept clean to match inference-time conditioning and otherwise noised to improve robustness to imperfect context. We also randomly drop the text condition to enable classifier-free guidance~\cite{ho2022classifierfreediffusionguidance}.

\textit{Action-conditioned video generation.} Since large-scale video data contain limited and heterogeneous action annotations, we first infer a unified action representation directly from video. Specifically, we adopt CD-LAM~\cite{cdlam}, a causally debiased latent action model to encode observations into latent actions that primarily capture embodiment dynamics. These latent actions are injected through AdaLN, allowing the model to learn action-conditioned dynamics using the same large-scale video mixture. 
%To support real robot control, we further learn a lightweight MLP layer that maps embodiment-specific actions $\mathbf{a}_t$ into the latent-action space of CD-LAM. The bridge and world model are then fine-tuned jointly on target embodiment data, yielding an action-conditioned predictor

\subsubsection{Stage 2: Mid-training with Causal Chain-of-thought}
The second stage teaches CausalWM to progressively predict intermediate variables and use them as context for future video generation. We construct the causal CoT with three commonly used visual features: \textbf{optical flow}, which captures scene motion; \textbf{depth}, which describes scene geometry; and \textbf{pointmaps}, which further lift depth into 3D spatial structure. Together, these features characterize complementary aspects of physical evolution from motion to geometry and can all be automatically extracted from raw videos using off-the-shelf models, making the supervision scalable to large-scale unlabeled video data. We encode these intermediate features with the same frozen VAE used for RGB videos and arrange them as an ordered reasoning sequence before the future observation. During training, we randomly select the $k$-th variable as the prediction target: all preceding variables are provided, the current variable is corrupted with diffusion noise and optimized using the flow-matching objective, while subsequent variables are masked from the current prediction:
\begin{equation}
\mathcal{L}_{\mathrm{CoT}}
=
\mathbb{E}_{k,\sigma,\boldsymbol{\epsilon}}
\left[
\left\|
\boldsymbol{v}_{\theta}
\!\left(
\hat{\boldsymbol{c}}_{r_k}^{\sigma}
\,\middle|\,
\boldsymbol{z}_{1:t},
\hat{\boldsymbol{c}}_{r_{<k}},
\boldsymbol{c},
\sigma
\right)
-
\boldsymbol{v}_{r_k,\sigma}
\right\|_2^2
\right],
\end{equation}
where $\hat{\boldsymbol{c}}_{r_{<k}}$ contains the preceding variables in the causal chain of thought, and $\boldsymbol{v}_{r_k,\sigma}$ denotes the target denoising velocity for the $k$-th variable. During inference, preceding variables are removed and replaced by the model's own outputs, which are generated sequentially and reused as context until the final future video is produced.

\subsubsection{Stage 3: Post-training with Multi-objective RL}
Future video generation is inherently open-ended: multiple plausible futures may exist for the same context, while supervised objectives alone do not directly optimize whether a complete generation is physically consistent, visually coherent, and task-relevant. We therefore further post-train CausalWM with reinforcement learning to directly improve the quality of generated futures.
Specifically, we adopt DiffusionNFT~\cite{diffusionnft}, which performs group-based policy optimization in a manner similar to GRPO~\cite{flowgrpo,deepseekmath_grpo,dancegrpo}. For each input context, the model samples a group of candidate future videos, which are evaluated by a set of reward functions. We consider complementary objectives including physical consistency, temporal coherence, visual quality, and task completion, and aggregate them into a unified reward for optimization. The resulting scores are combined into one reward, normalized within the group, and mapped to a preference weight $r\in[0,1]$. Each sampled future is re-noised to $\boldsymbol{x}_t$ with the observation held clean, and the model is updated with

\begin{equation}
\mathcal{L}_{RL}=\mathbb{E}_{\boldsymbol{c},\,\pi^{\mathrm{old}}(\boldsymbol{z}_0\mid\boldsymbol{c}),\,t}\Big[\,r\,\big\|\boldsymbol{v}^{+}_\theta(\boldsymbol{z}_{1:t},\boldsymbol{c},t)-\boldsymbol{v}\big\|_2^2+(1-r)\,\big\|\boldsymbol{v}^{-}_\theta(\boldsymbol{z}_{1:t},\boldsymbol{c},t)-\boldsymbol{v}\big\|_2^2\Big],
\end{equation}

\begin{equation}
\begin{aligned}
\boldsymbol{v}^{+}_\theta(\boldsymbol{z}_{1:t},\boldsymbol{c},t)&:=(1-\beta)\,\boldsymbol{v}^{\mathrm{old}}(\boldsymbol{z}_{1:t},\boldsymbol{c},t)+\beta\,\boldsymbol{v}_\theta(\boldsymbol{z}_{1:t},\boldsymbol{c},t),\\
\boldsymbol{v}^{-}_\theta(\boldsymbol{z}_{1:t},\boldsymbol{c},t)&:=(1+\beta)\,\boldsymbol{v}^{\mathrm{old}}(\boldsymbol{z}_{1:t},\boldsymbol{c},t)-\beta\,\boldsymbol{v}_\theta(\boldsymbol{z}_{1:t},\boldsymbol{c},t).
\end{aligned}
\end{equation}
where $\boldsymbol{v}^{\mathrm{old}}$ is the model before the update, $\boldsymbol{v}^{+}_\theta$ and $\boldsymbol{v}^{-}_\theta$ are the implicit positive and negative branches as dual directions for policy optimization, and $\beta$ sets how far the update may move from $\boldsymbol{v}^{\mathrm{old}}$. High-reward futures pull the velocity toward themselves and low-reward futures push it away. Unlike supervised training, this RL stage allows the model to explore multiple possible reasoning trajectories and future outcomes rather than imitating a single target. In particular, the causal CoT can be jointly explored according to the quality of the final generated video, encouraging intermediate reasoning paths that lead to better physical dynamics and more accurate future predictions.

\subsection{Flexible Control Guidance through Causal CoT}
Beyond improving structured physical reasoning, causal CoT also enhances the model's ability to exploit in-context visual features, providing a flexible interface for adding new control signals beyond those used during training. Since intermediate variables are repeatedly inserted into the shared token sequence and reused to guide later predictions, CausalWM learns to attend to informative visual context in a general way. Accordingly, any control signal that can be represented as compatible visual representations can be introduced as an additional in-context feature without modifying the backbone. For example, action trajectory videos rendered by a simulator can be used to replace an existing intermediate stream. Through fine-tuning on few episodes, the model can then learn to attend to these in-context controls and generate future videos that remain consistent with the specified trajectories. The same way can naturally extend to other controls such as target poses, object tracks, or geometric constraints.

 % Ziming & Shuang

% Experiments
\section{Experiments}
\label{sec:experiments}

\subsection{Experimental Setup}
\label{sec:experimental-setup}

\paragraph{Benchmarks and Metrics.}
We evaluate CausalWM (CWM) on two public benchmarks covering action-conditioned multi-view prediction and language-conditioned video generation. Unless otherwise noted, baseline results are taken from the corresponding official leaderboards, and our submissions follow the official evaluation pipelines.

\par\medskip
\noindent\textit{TriWorldBench~\cite{triworldbench2026}.}
TriWorldBench targets action-conditioned generation for a three-camera manipulation setup, with synchronized observations from a head camera and two wrist cameras. We evaluate on the full 500-episode test set spanning 50 manipulation tasks. The aggregate TWB-Score averages 19 metrics across six dimensions: tri-view consistency, task alignment, physical and 3D coherence, motion quality, temporal consistency and visual quality. We compare CWM with baselines that have publicly available technical reports, using the official leaderboard snapshot of Sep. 11, 2026 (36 models); metric rankings in Table~\ref{tab:triworldbench-full} are computed only among the models shown.

\par\medskip
\noindent\textit{PAI-Bench~\cite{paibench}.}
PAI-Bench-G is the video-generation track of PAI-Bench, containing 1,044 video--prompt pairs for evaluating physical-world prediction. We evaluate on its robot domain, comprising 174 prompts with 913 binary VQA questions in total (3--14 per prompt). For each prompt we generate five videos with different random seeds. Each question is scored by a Qwen3-VL-235B-A22B-Instruct~\cite{qwen3vl} judge; the Domain Score aggregates accuracy with equal weight per video and is reported on a 0--100 scale.

% \par\medskip
% \noindent\textit{RBench~\cite{data_rovidx}.}
% RBench contains 650 image--text pairs organized by five task categories and four robot types. We evaluate on the single-arm and dual-arm embodiments (100 cases each), generating one video per case with a fixed per-case seed. A Qwen3-VL-235B-A22B-Instruct judge scores physical-semantic plausibility, task adherence and robot-subject stability, while motion amplitude and motion smoothness are computed with dedicated motion metrics. Scores are aggregated within each embodiment, and the final score averages the task-completion and visual-quality components on a 0--1 scale.

\paragraph{Implementation Details.}
All models are built on the LTX-2.3-22B video transformer~\cite{ltx-video,ltx2} with a Gemma-3-12B text encoder~\cite{gemma3}, while the audio part parameters are removed. Video VAE and text-conditioning modules are kept frozen throughout. The VAE produces 128-channel latents with a temporal stride of $8$ and spatial strides of $32\times32$. We use AdamW~\cite{adamw} with $(\beta_1,\beta_2)=(0.9,0.999)$, weight decay $0.01$ and $\epsilon=10^{-8}$, gradient clipping at a norm of $1.0$, bfloat16 precision and gradient checkpointing. Noise levels follow the backbone's sequence-length-dependent shifted logit-normal sampler with a $10\%$ uniform mixture. %Below we describe the two training stages of Section~\ref{sec:multi-stage-training}; the checkpoint used for each benchmark is listed at the end.

\par\medskip
\noindent\textit{Stage 1: Pre-training.}
Starting from LTX-2.3-22B, we first perform language-conditioned pre-training for 60,000 updates on 128 H100 GPUs using 65-frame single-view clips at $640\times480$ resolution and an effective batch size of 256. We sample one to eight latent history frames, keep the history clean with probability $0.5$, and apply a caption-drop rate of $0.2$. From this checkpoint, we further train an action-conditioned variant for 30,000 updates on 32 H200 GPUs using 32-dimensional CD-LAM~\cite{cdlam} latent actions, with eight transition codes concatenated per latent frame and injected through AdaLN. We also train a multi-view variant for 120,000 updates on 32 H200 GPUs by horizontally concatenating two to four synchronized camera views and appending their order to the caption, while otherwise following the single-view training configuration.

\par\medskip
\noindent\textit{Stage 2: Causal CoT mid-training.}
Starting from the language-conditioned checkpoint, we train CausalWM for 50,000 updates on 64 H200 GPUs using 121-frame clips at $640\times480$. RGB, optical flow, and pointmaps use modality-specific input/output projections and AdaLN embeddings, with the projections initialized from the pretrained RGB branch. Each update uniformly samples one stage as the prediction target and provides preceding CoT variables as clean context. Optical-flow targets are extracted with SEA-RAFT~\cite{sea-raft}, while depth and camera intrinsics are estimated with VGGT-Omega-1B-512~\cite{vggt-omega} and converted into normalized XYZ pointmaps; both streams are encoded offline with the frozen VAE and cached. For three-view action-conditioned control on TriWorldBench, we additionally render joint and gripper trajectories from the robot URDF using forward kinematics from the head and two wrist cameras, concatenate the three views, and encode them as visual in-context controls. Starting from the pretrained three-view action-conditioned model, we fine-tune the transformer, action bridge, and control projection for 34,000 updates, using 129-frame clips with one clean anchor frame, eight history frames, and 120 target frames.

\paragraph{Evaluation Protocol.}
For PAI-Bench % and RBench
we use the single-view CoT model and generate 121-frame videos at $640\times480$ from a single observation and the language instruction, with flow and pointmaps predicted sequentially before RGB. On PAI-Bench we use 4 denoising steps per CoT stage without classifier-free guidance, and report the mean over five seeds per prompt.
% On RBench we use 10 denoising steps per stage without classifier-free guidance, generating one video per case with a fixed seed.
For TriWorldBench we use the three-view action-conditioned model and generate videos matching the length of the supplied action trajectory in an autoregressive manner: each 129-frame window is conditioned on nine frames (one episode-anchor frame and eight recent history frames) and predicts the next 120 frames at $1920\times480$, using 30 denoising steps without classifier-free guidance.
\subsection{Main Results}
\label{sec:main-results}

\paragraph{Multi-view Action-Conditioned Embodied World Model.}

Table~\ref{tab:triworldbench-full} compares CWM with eight baselines from the official TriWorldBench leaderboard~\cite{triworldbench2026}. All compared baselines have publicly available websites or technical reports. We select seven key metrics covering all six official evaluation dimensions for a compact assessment of embodied prediction. CWM achieves the highest TWB-Score of 66.04, exceeding the strongest baseline in the table, BWM (65.54), by 0.50 points. Among the compared models, CWM ranks first in VLM Consistency (averaged over I--III), VQA Consistency, Instruction Following, Perspective, and Image Quality, and second in Trajectory Accuracy and Subject Consistency. The complete leaderboard with all 19 metrics is provided in Appendix~\ref{app:triworldbench-complete}.
\begin{table}[htbp]
  \centering
  \caption{TriWorldBench leaderboard results snapshot: Sep. 11, 2026. TWB-Score is computed by weighted sum of all 19 evaluation metrics, and we select 7 key measures from it. VLM Consistency averages the three official VLM-as-judge metrics, and Cons. denotes Consistency. We select all compared baselines that have official websites or technical reports. Bold and underlined scores denote first and second place, respectively.}
  \label{tab:triworldbench-full}
  \renewcommand{\baselinestretch}{1}
  \fontsize{8.8}{11}\selectfont
  \setlength{\tabcolsep}{2pt}
  \renewcommand{\arraystretch}{1.3}
  \begin{tabularx}{\linewidth}{@{}>{\raggedright\arraybackslash}p{0.205\linewidth}*{7}{>{\centering\arraybackslash}X}!{\vrule width 0.5pt}>{\centering\arraybackslash}p{0.08\linewidth}@{}}
    \toprule[0.7pt]
    \textbf{Model}
    & {\fontsize{8}{9.5}\selectfont\bfseries\shortstack{VLM\\Cons.}}
    & {\fontsize{8}{9.5}\selectfont\bfseries\shortstack{VQA\\Cons.}}
    & {\fontsize{8}{9.5}\selectfont\bfseries\shortstack{Instruction\\Following}}
    & {\fontsize{8}{9.5}\selectfont\bfseries\raisebox{0.5\baselineskip}{Perspective}}
    & {\fontsize{8}{9.5}\selectfont\bfseries\shortstack{Trajectory\\Accuracy}}
    & {\fontsize{8}{9.5}\selectfont\bfseries\shortstack{Subject\\Cons.}}
    & {\fontsize{8}{9.5}\selectfont\bfseries\shortstack{Image\\Quality}}
    & \cellcolor{blue!18}{\fontsize{9}{11}\selectfont\bfseries\shortstack{TWB-\\Score}} \\
    \midrule[0.35pt]
    Ctrl-World~\cite{ctrl-world} & 64.20 & 40.90 & 38.90 & 45.15 & 6.09 & 77.05 & 18.37 & \cellcolor{blue!7}{\fontsize{9}{11}\selectfont 38.98} \\
    Genie Envisioner~\cite{genie-envisioner} & 68.15 & 36.72 & 20.20 & \underline{87.91} & 0.84 & \textbf{91.40} & 19.36 & \cellcolor{blue!7}{\fontsize{9}{11}\selectfont 40.73} \\
    Motus~\cite{motus} & 72.82 & 45.33 & 52.28 & 44.20 & 13.77 & 77.37 & 16.08 & \cellcolor{blue!7}{\fontsize{9}{11}\selectfont 42.35} \\
    DreamDojo~\cite{dreamdojo} & 79.53 & 47.82 & 49.72 & 59.39 & 16.61 & 77.37 & 24.45 & \cellcolor{blue!7}{\fontsize{9}{11}\selectfont 51.72} \\
    Fysiverse-Video~\cite{fysiverse-video} & 87.66 & 63.32 & 70.18 & 83.86 & 37.40 & 83.65 & 36.17 & \cellcolor{blue!7}{\fontsize{9}{11}\selectfont 63.62} \\
    XiaomiAutoWM~\cite{xiaomi-auto-worldmodel} & 86.88 & \underline{69.27} & 66.40 & 78.28 & 42.85 & 83.61 & 33.63 & \cellcolor{blue!7}{\fontsize{9}{11}\selectfont 63.94} \\
    WoVR Plus~\cite{wovr} & 87.52 & 69.11 & 70.67 & 85.49 & \textbf{45.27} & 84.22 & 36.22 & \cellcolor{blue!7}{\fontsize{9}{11}\selectfont 65.39} \\
    BWM~\cite{bwm2026} & \underline{88.12} & 65.94 & \underline{73.84} & 87.20 & 41.97 & 84.20 & \underline{38.71} & \cellcolor{blue!7}{\fontsize{9}{11}\selectfont \underline{65.54}} \\
    \midrule[0.35pt]
    \rowcolor{opensourcebg}
    \textbf{CWM} (Ours) & \textbf{88.63} & \textbf{70.03} & \textbf{76.78} & \textbf{91.16} & \underline{42.98} & \underline{84.55} & \textbf{43.24} & \cellcolor{blue!18}{\fontsize{9}{11}\selectfont \textbf{66.04} } \\
    \bottomrule[0.7pt]
  \end{tabularx}
  \par\vspace{4pt}
  %{\fontsize{8.2}{10.2}\selectfont\raggedright Official leaderboard snapshot: Sep. 11, 2026.\par}
\end{table}

\paragraph{Language-Conditioned Embodied World Model}

% Tables~\ref{tab:paibench-results} and~\ref{tab:rbench-results} compare CWM with the leaderboards. [CWM results and discussion to be filled in.]
Table~\ref{tab:paibench-results} compares CWM with baseline models on the robot (RO) domain of PAI-Bench-G. CWM and Cosmos3-Super~\cite{cosmos3} are evaluated locally. For CWM, we use rewritten versions of the original PAI-Bench-G prompts designed to better match the style of the pretraining captions, while Cosmos3-Super uses the same inference settings reported in the Cosmos 3 technical report.

Consistent with the Cosmos 3 technical report, we were unable to reproduce the PAI-Bench-G leaderboard scores exactly. CWM achieves an RO score of 89.9, the highest among the models compared in Table~\ref{tab:paibench-results}, while Cosmos3-Super scores 89.7. Despite the discrepancy in absolute scores, the relative score comparison remains informative for assessing CWM's performance against the baselines.

\begin{table}[!htbp]
  \centering
  \caption{Language-conditioned generation results on the robot domain (RO) of PAI-Bench-G. Scores are reported on a 0--100 scale as mean of per-video Visual Question Answering (VQA) accuracy in the robot domain. CausalWM (CWM) and Cosmos3-Super are evaluated locally using official evaluation protocol with Qwen3-VL-235B-A22B-Instruct~\cite{qwen3vl} as judge, while results for the other baselines are taken from the official leaderboard. Detailed RO subcategory scores for the two locally-evaluated models are provided in Appendix~\ref{app:paibench-g-details}. }
  \label{tab:paibench-results}
  \renewcommand{\baselinestretch}{1}
  \fontsize{8.8}{11}\selectfont
  \setlength{\tabcolsep}{6pt}
  \renewcommand{\arraystretch}{1.3}
  \begin{minipage}{0.86\linewidth}
  \begin{tabularx}{\linewidth}{@{\hspace{6pt}}
    >{\raggedright\arraybackslash}X!{\vrule width 0.5pt}
    >{\centering\arraybackslash}p{0.24\linewidth}@{\hspace{6pt}}}
    \toprule[0.7pt]
    {\fontsize{9.5}{11}\selectfont\bfseries Model} & \cellcolor{blue!18}{\fontsize{9.5}{11}\selectfont\bfseries RO $\uparrow$} \\
    \midrule[0.35pt]
    % Source: https://huggingface.co/spaces/shi-labs/physical-ai-bench-leaderboard (Generation, Robot).
    % Cosmos3-Super uses local evaluation results; other baseline scores are from the official leaderboard.
    % Rows within each group are ordered by ascending RO score.

    \rowcolor{gray!12}
    \multicolumn{2}{@{\hspace{6pt}}l@{\hspace{6pt}}}{\fontsize{9.5}{11}\selectfont\bfseries Commercial Models} \\
    Veo-3~\cite{veo3} & \cellcolor{blue!7}{\fontsize{9}{11}\selectfont 86.9} \\
    \midrule[0.35pt]

    \rowcolor{gray!12}
    \multicolumn{2}{@{\hspace{6pt}}l@{\hspace{6pt}}}{\fontsize{9.5}{11}\selectfont\bfseries Open-source Models} \\
    LTX-Video-13B~\cite{ltx-video} & \cellcolor{blue!7}{\fontsize{9}{11}\selectfont 70.1} \\
    CogVideoX-5b-I2V~\cite{cogvideox} & \cellcolor{blue!7}{\fontsize{9}{11}\selectfont 74.0} \\
    Wan2.2-TI2V-5B~\cite{wan2-2} & \cellcolor{blue!7}{\fontsize{9}{11}\selectfont 79.3} \\
    Cosmos-Predict2.5-14B~\cite{cosmos2-5} & \cellcolor{blue!7}{\fontsize{9}{11}\selectfont 79.9} \\
    Wan2.1-I2V-14B-720P~\cite{wan2-2} & \cellcolor{blue!7}{\fontsize{9}{11}\selectfont 80.1} \\
    Wan2.2-I2V-A14B~\cite{wan2-2} & \cellcolor{blue!7}{\fontsize{9}{11}\selectfont 81.7} \\
    Cosmos3-Super~\cite{cosmos3} & \cellcolor{blue!7}{\fontsize{9}{11}\selectfont \underline{89.7}} \\
    \midrule[0.35pt]
    \rowcolor{opensourcebg}
    \textbf{CWM} (Ours) & \cellcolor{blue!18}{\fontsize{9}{11}\selectfont \textbf{89.9}} \\
    \bottomrule[0.7pt]
  \end{tabularx}
  \end{minipage}

\end{table}

\FloatBarrier

\subsection{Effect of Causal Chain-of-thought}

\paragraph{Support Extremely Few Denoising Steps.}
Causal CoT offers a \emph{space--time trade-off} for future prediction. Here, the additional space is the token context allocated to intermediate physical variables. Explicit motion and geometry features provide structured conditions for later predictions, which can reduce the need for repeated denoising refinement. This richer context incurs additional storage and attention costs, but can support a smaller denoising budget. We examine the few-step behavior of this design by varying the number of denoising iterations while keeping the CoT structure fixed.

We evaluate CausalWM only on the robot domain (RO) of PAI-Bench, using Qwen3-VL-235B-A22B-Instruct as the judge, consistent with the main results in Table~\ref{tab:paibench-results}. We progressively reduce the denoising budget from 20 to 1 step per generation stage. Table~\ref{tab:paibench-few-step} reports the RO scores on a 0--100 scale and generation speedups relative to 20/20/20 under otherwise fixed settings. Reducing the schedule from 20/20/20 to 4/4/4 increases the score from 86.54 to 88.86 while reducing generation time from 76.43 to 24.25 seconds, a $3.15\times$ speedup. With only one denoising step per stage, the 1/1/1 schedule achieves an RO score of 88.84 in 14.81 seconds, yielding a $5.16\times$ speedup over 20/20/20. This score is only 0.02 points below the best observed score and 2.30 points above the 20/20/20 schedule. These results show that CausalWM retains strong performance with a single denoising step per stage---three denoising steps across the complete causal chain---consistent with the intended space--time trade-off.

\begin{table}[!htbp]
  \centering
  \caption{\textbf{Few-step generation on the robot domain of PAI-Bench.} Only the robot-domain (RO) score is evaluated, using Qwen3-VL-235B-A22B-Instruct as the judge. Scores are on a 0--100 scale. Each schedule is evaluated on 174 tasks with five random seeds per task (870 generations). The three entries specify the denoising budgets of the successive generation stages. Speedup is computed relative to 20/20/20.}
  \label{tab:paibench-few-step}
  \small
  \renewcommand{\arraystretch}{1.12}
  \begin{tabular*}{0.78\textwidth}{@{\extracolsep{\fill}}ccc@{}}
    \toprule
    Denoising steps & RO score $\uparrow$ & Speedup $\uparrow$ \\
    \midrule
    20/20/20 & 86.54 & $1.00\times$ \\
    10/10/10 & 86.75 & $1.74\times$ \\
    8/8/8   & 87.53 & $2.06\times$ \\
    4/4/4   & \textbf{88.86} & $3.15\times$ \\
    2/2/2   & 88.71 & $4.41\times$ \\
    1/1/1   & 88.84 & $\mathbf{5.16}\times$ \\
    \bottomrule
  \end{tabular*}
\end{table}

\begin{figure}[!htbp]
  \centering
  \includegraphics[width=0.96\linewidth]{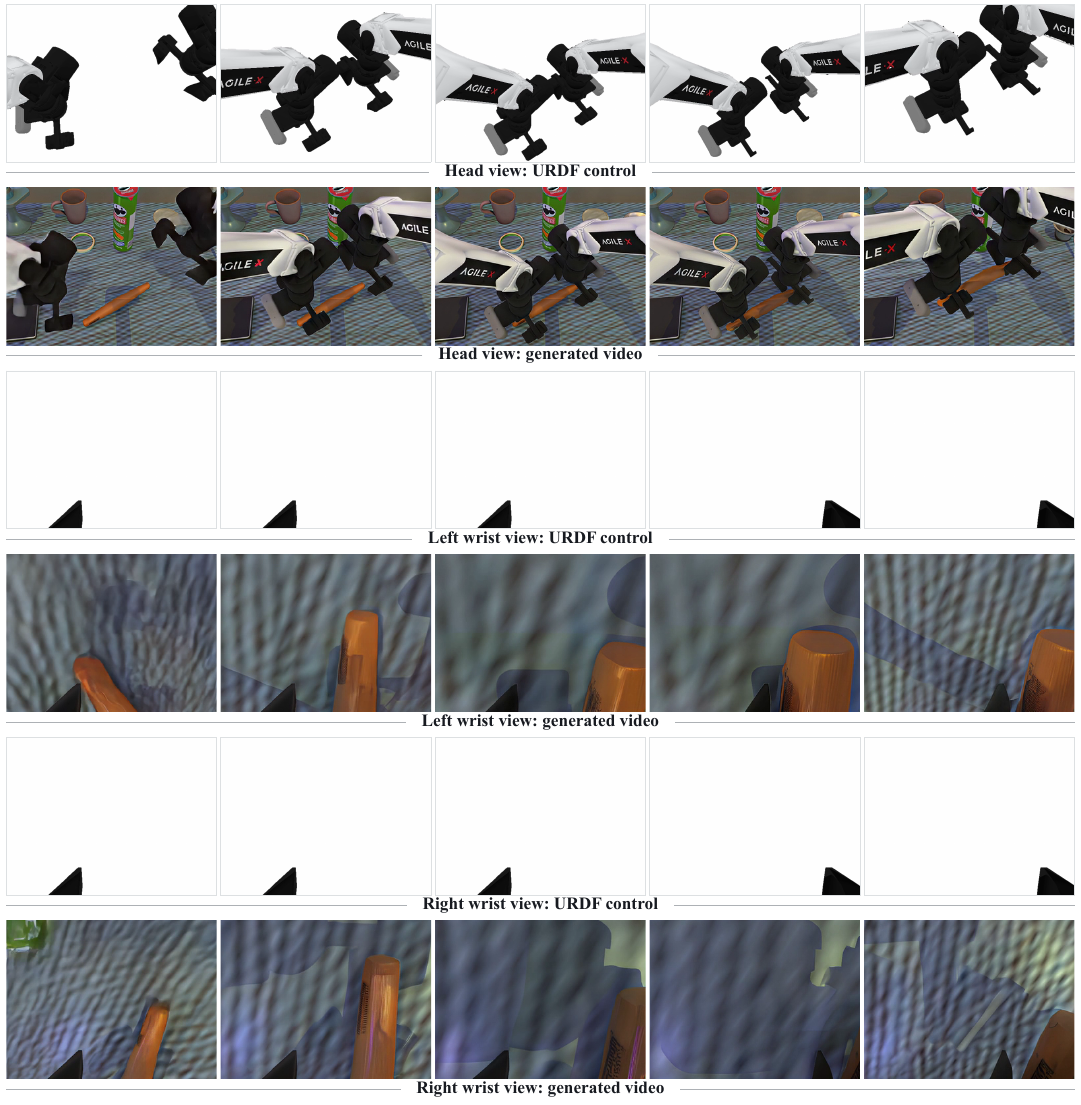}
  \caption{\textbf{Action guidance through visual CoT on TriWorldBench.} Each pair of rows shows the supplied URDF control video and the generated RGB video for the head, left wrist and right wrist views, respectively. Columns show synchronized keyframes in chronological order. The rendered robot motion serves as an external visual condition within causal CoT.}
  \label{fig:triworld-control}
\end{figure}

\paragraph{Support In-context Feature Guidance.}We illustrate the visual control interface of causal CoT with a TriWorldBench case (Figure~\ref{fig:triworld-control}). We convert the supplied action trajectory into synchronized control videos by rendering the robot's URDF model from the head and two wrist cameras. These videos serve as an externally supplied intermediate stream within causal CoT, conditioning future RGB generation together with the initial visual observation. After fine-tuning with this representation, CausalWM generates three-view videos that follow the prescribed robot motion. In the illustrated sequence, the generated arms approach and lift the object in correspondence with the rendered trajectory, while the wrist views show the interaction from the moving cameras. This case illustrates how actions can be expressed as visual conditioning signals and incorporated into the CoT interface for controllable video generation.

\FloatBarrier

\subsection{Case Study}
\label{sec:case-study}

\begin{figure}[!htbp]
  \centering
  \includegraphics[width=0.96\linewidth]{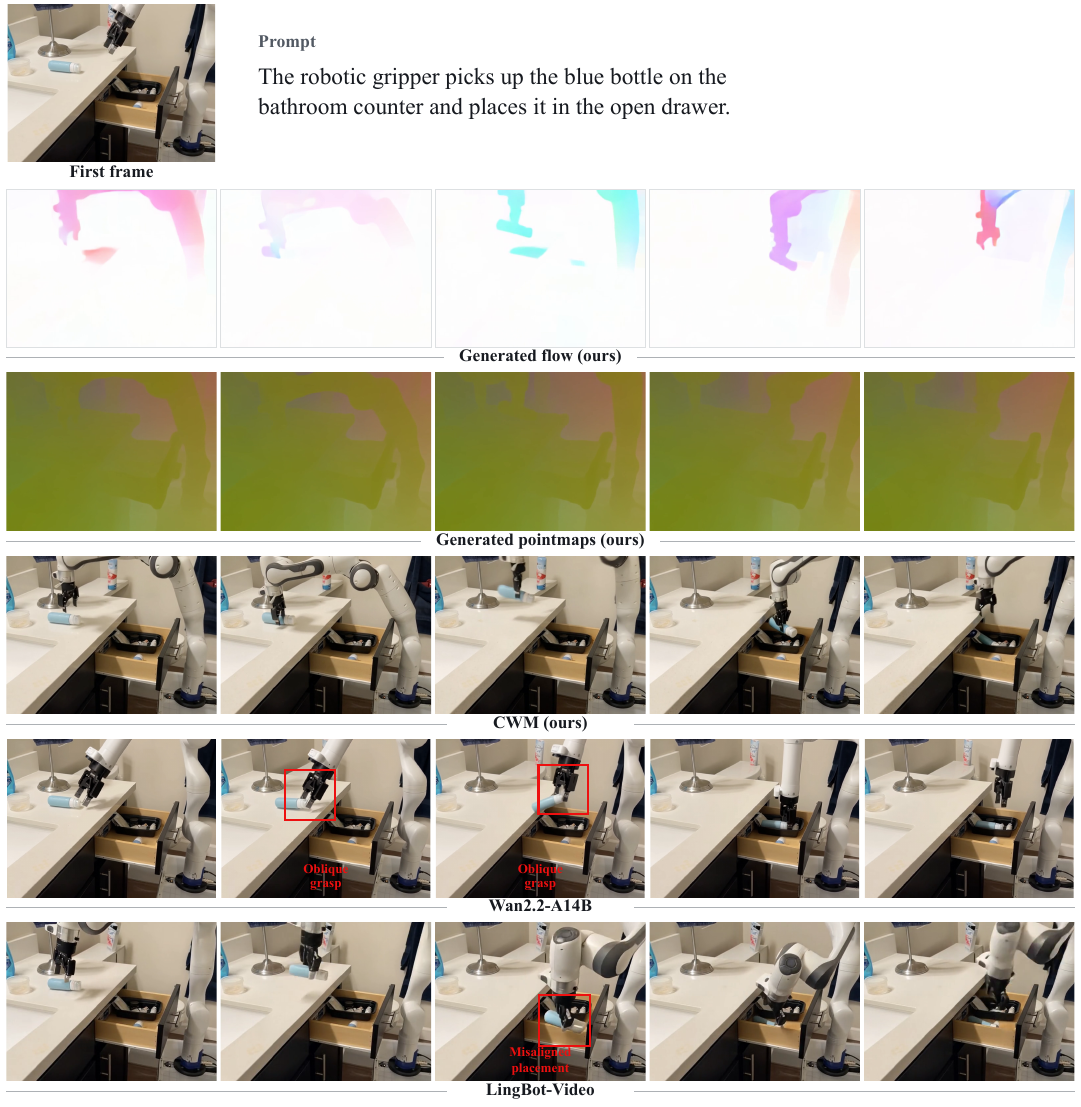}
  \caption{\textbf{Motion, geometry and RGB predictions on a bottle-to-drawer task.} The first row shows the input observation and instruction. The next three rows show CWM's generated flow, pointmaps and RGB, aligned at the same frame indices. The final two rows show Wan2.2-A14B and LingBot-Video. Red boxes mark oblique gripper contact and transient bottle misplacement, respectively. Keyframes are selected independently for each model and arranged chronologically. Baseline keyframes are nonuniformly spaced, and columns are not temporally synchronized across models.}
  \label{fig:case-bottle-drawer}
\end{figure}

\begin{figure}[!htbp]
  \centering
  \includegraphics[width=0.96\linewidth]{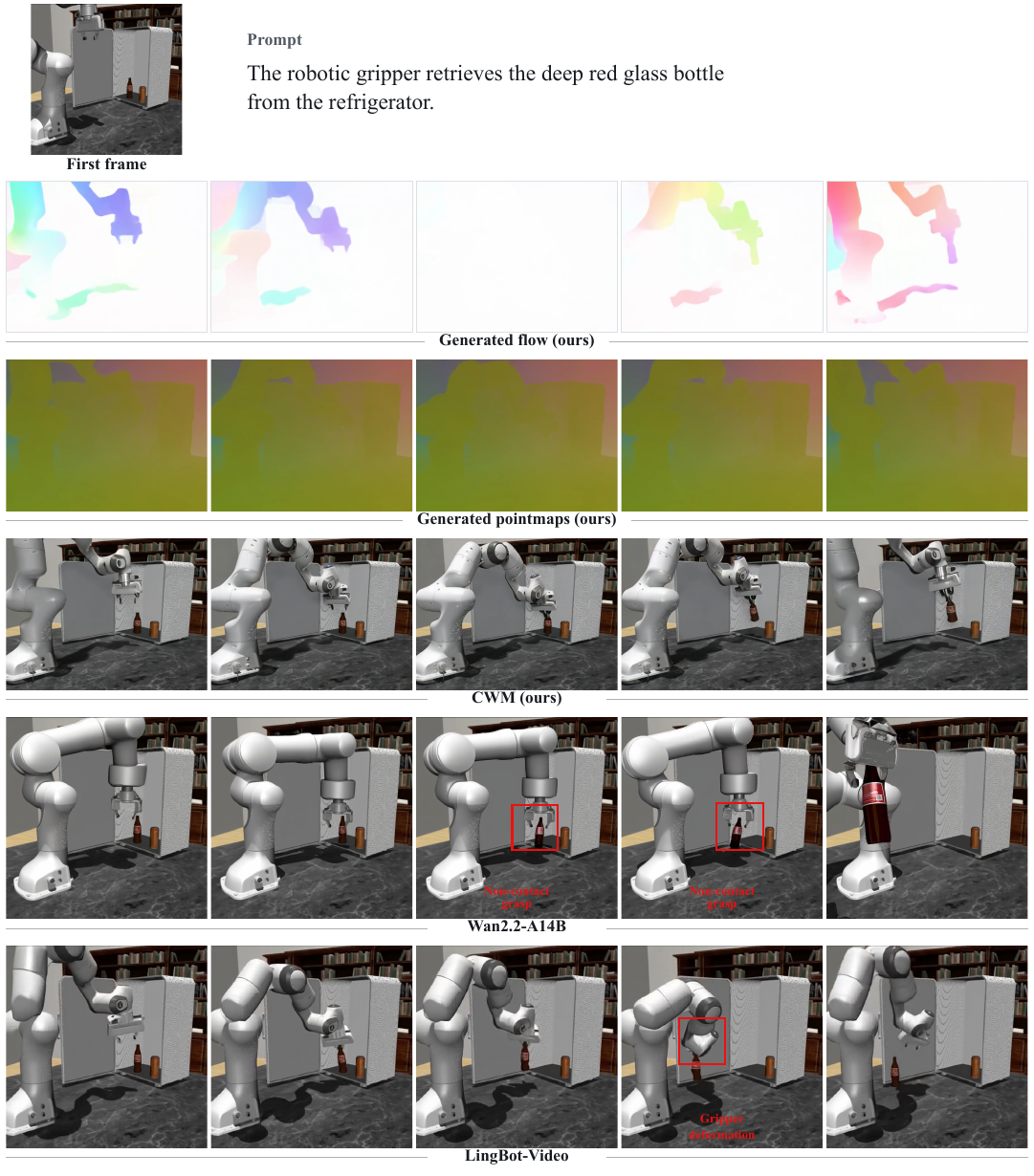}
  \caption{\textbf{Bottle retrieval.} The first row shows the initial observation and instruction, followed by CWM's generated flow, pointmaps and RGB, and the RGB predictions of Wan2.2-A14B and LingBot-Video. Red boxes highlight non-contact grasping and gripper deformation, respectively. Keyframes are selected independently for each model and arranged chronologically; columns are synchronized only across CWM's three generated streams.}
  \label{fig:case-bottle-retrieval}
\end{figure}

\begin{figure}[!htbp]
  \centering
  \includegraphics[width=0.96\linewidth]{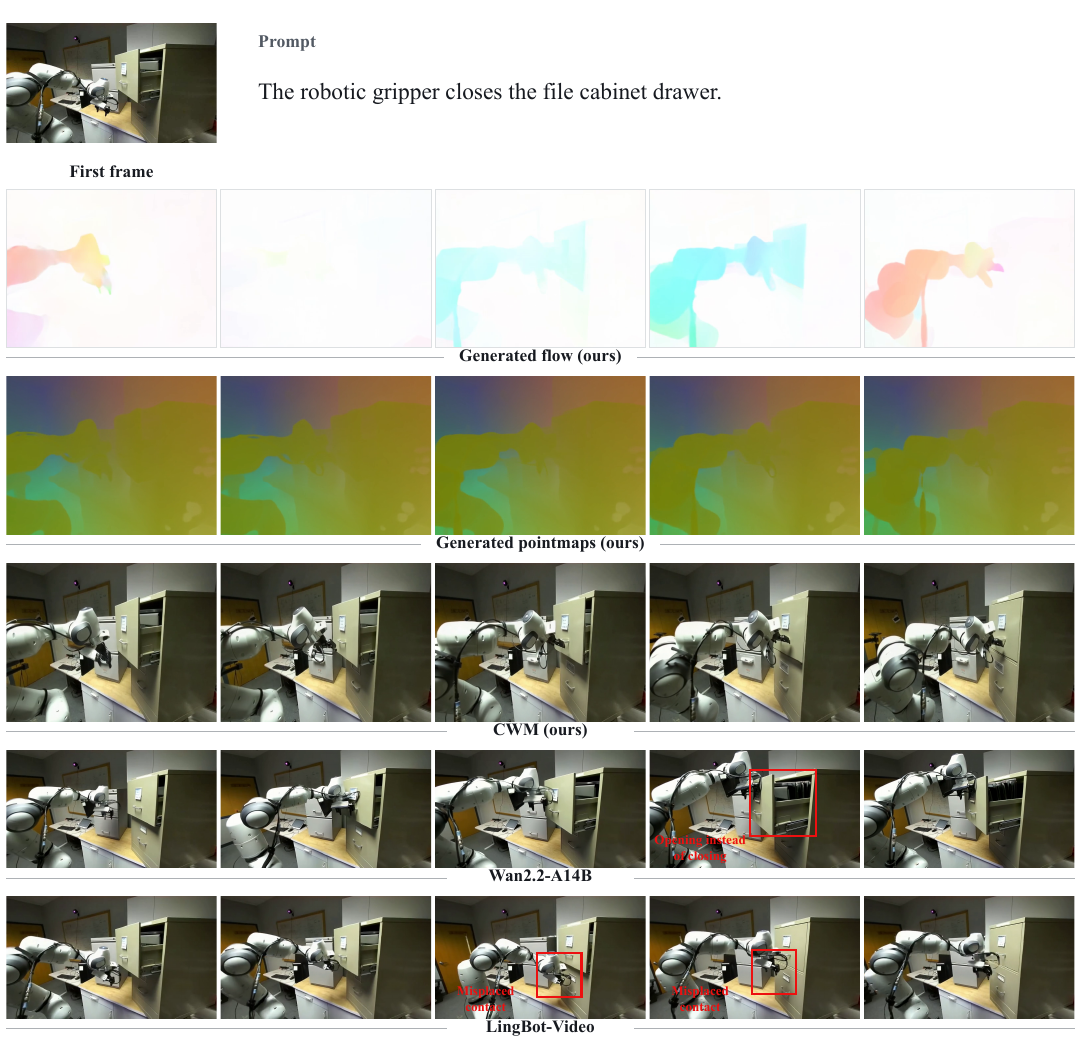}
  \caption{\textbf{Drawer closing.} The first row shows the initial observation and instruction, followed by CWM's generated flow, pointmaps and RGB, and the RGB predictions of Wan2.2-A14B and LingBot-Video. Red boxes highlight drawer opening contrary to the instruction and misplaced contact below the target drawer, respectively. Keyframes are selected independently for each model and arranged chronologically; columns are synchronized only across CWM's three generated streams.}
  \label{fig:case-drawer-closing}
\end{figure}

We examine grasping and placement in a language-conditioned bottle-to-drawer manipulation task (Figure~\ref{fig:case-bottle-drawer}). Baseline videos of Wan2.2-A14B and LingBot-Video~\cite{wan2-2,lingbot-video} are generated with their officially released checkpoints under default inference settings. CWM approaches the bottle with a near-vertical gripper, grasps its body and transfers it into the open drawer. Wan2.2-A14B instead grasps close to the cap with an oblique gripper pose. LingBot-Video initially positions the bottle across the drawer's front edge, with the cap extending beyond it, before lowering the bottle inside. CWM maintains better alignment between the bottle and the drawer during placement in this example. Two further language-conditioned cases illustrate differences in contact and instruction following. In bottle retrieval (Figure~\ref{fig:case-bottle-retrieval}), Wan2.2-A14B shows a bottle suspended below an open gripper, while LingBot-Video exhibits gripper deformation. In drawer closing (Figure~\ref{fig:case-drawer-closing}), Wan2.2-A14B pulls the drawer open, whereas LingBot-Video places the gripper below the target drawer. CWM grasps and lifts the bottle in the first case and pushes the target drawer closed in the second.

These observations are consistent with the complementary roles of motion and geometry prediction in our causal CoT. Predicted flow provides explicit motion context for coordinating the gripper and bottle during grasping and transfer, while predicted pointmaps provide geometric context for positioning the bottle relative to the drawer. Both predicted streams then condition future RGB generation.

\section{Related Work} % Ethan
\label{sec:related-work}

\paragraph{Embodied World Model.}
Embodied world models aim to predict how an environment evolves under an agent's control, providing a learned simulator for action selection, planning, and policy learning~\citep{ha2018world, hafner2023mastering}. Early approaches~\citep{ha2018world, hafner2019learning, hafner2019dream, hafner2020mastering, hafner2023mastering} roll out a compact latent state with a recurrent transition model. The policy model is trained completely in the latent state. This is efficient for control, but the latents throw away most visual detail and do not transfer well to open-world or language-specified tasks. Recent line treats the problem as conditional video generation: given the current frames and a control signal (e.g. low-level actions or a language instruction), the model predicts future observations directly in pixel space~\citep{yang2023learning, du2023learning}. Pre-trained on large video corpora, these models pick up rich dynamics, and the same recipe has since been pushed to robot manipulation~\citep{wu2024unleashing}, driving~\citep{gao2024vista}, and foundation-scale systems such as Cosmos~\citep{agarwal2025cosmos}. 
What these models share is that they learn one conditional distribution mapping context and control straight to future frames. The physical knowledge stays entangled in the latents, so it is hard to tell whether a model has actually captured the causal variables behind an interaction or has just found shortcut correlations that fit the training loss~\citep{geirhos2020shortcut}. Some methods add optical flow or correspondences as an auxiliary signal to supply motion cues~\citep{ko2024learning}, but this is usually a single-side input rather than a set of variables organized in causal order. We instead break future prediction into a causal chain of intermediate physical variables (optical flow and pointmap), which makes each step explicit and supervisable and pushes the model to learn the causal structure.

\paragraph{Chain-of-thought Reasoning.} Chain-of-thought prompting was introduced for large language models, where a model solves a hard problem by first writing out intermediate reasoning steps instead of jumping straight to the answer~\citep{wei2022-llm-cot}. This simple change gives a large boost on tasks like arithmetic and multi-step question answering, and later work showed the steps can be produced without hand-written examples~\citep{kojima2022large} and made more reliable by sampling several chains and taking the majority answer~\citep{wang2022self}. The common thread is that breaking a prediction into ordered intermediate steps is easier to learn and to get right than predicting the final answer in one shot.
More recent work carries this idea beyond text. Multimodal models attach reasoning chains to images for visual question answering~\citep{zhang2023multimodal,shao2026learning}, and a few image-generation methods~\citep{feng2023layoutgpt, lian2024llmgrounded} first predict an intermediate plan, such as a layout, before rendering the final picture. What these share with the language case is the order: an explicit intermediate result is produced first and then used to constrain what comes next. We take the same view for embodied prediction. Instead of predicting future pixels directly, we generate physically meaningful intermediates in sequence, each step conditioned on the ones before it, and produce the future frames from all of them.

\paragraph{Causal Representation and Reasoning.} Causal representation captures the underlying causal factors of a system, together with the relations among them, rather than settling for entangled features that merely fit the data~\citep{scholkopf2021toward}. Theoretically, the same observations can be represented by many different sets of latent factors, so fitting the data alone cannot tell which one corresponds to the true causal factors. This line of work introduces additional structure, such as interventions, temporal ordering, or known mechanisms, under which the latent variables can be tied back to real causal factors~\citep{scholkopf2021toward, khemakhem2020variational}. We do not attempt to identify unknown causal factors from data. A representative method in the temporal and interactive setting closest to our idea learns latent variables whose transitions follow a causal graph, using action or time as a weak supervisory signal to separate the factors that drive dynamics~\citep{lippe2022citris}. We call this causal reasoning. Specifically, we fix a small set of physically meaningful variables such as optical flow and pointmap and impose a causal generation order over them, so that future frames are produced through, and constrained by, these intermediate steps rather than in a single entangled mapping.

\section{Conclusion}
In this work, we introduced CausalWM, an embodied world model that formulates future prediction through causal chain-of-thought reasoning. Instead of relying solely on implicit physical knowledge encoded in latent representations, CausalWM organizes useful physical variables into explicit intermediate reasoning steps, providing structured supervision for modeling how physical states evolve under control signals. Built on 20K hours of diverse embodied data, our three-stage training paradigm combines large-scale video pre-training, causal CoT mid-training, and multi-objective RLVR post-training. CausalWM further exhibits emergent in-context reasoning capabilities beyond the explicitly supervised variables, enabling in-context visual feature guidance and efficient few-step generation, while achieving state-of-the-art performance across diverse embodied world-model benchmarks.

Looking forward, an important direction is to develop more general and expressive causal representation learning methods that can discover useful causal variables and their dependencies directly from large-scale interaction data, rather than relying on a predefined set of intermediate variables. Another promising direction is to extend CausalWM toward a unified world-action model, jointly modeling future world dynamics and executable actions within the same causal reasoning framework. Such a model could provide a tighter connection between physical understanding, prediction, and control, and ultimately support more capable long-horizon embodied intelligence.

\section*{Limitations}
CausalWM still has several limitations. First, our causal CoT relies on a predefined set of intermediate physical variables, which may not fully capture the causal structure required for more complex environments. Second, although CausalWM shows strong performance across multiple benchmarks, its generalization to substantially longer-horizon interactions and highly out-of-distribution physical settings remains to be further explored. Finally, the current model focuses primarily on world prediction; integrating action generation into a unified world-action modeling framework is an important direction for future work.

\clearpage
\phantomsection
\addcontentsline{toc}{section}{References}
\bibliographystyle{plainnat}
\bibliography{reference}

\beginappendix
\setcounter{figure}{0}
\setcounter{table}{0}
\renewcommand{\thefigure}{A\arabic{figure}}
\renewcommand{\thetable}{A\arabic{table}}

% TriWorldBench

\section{Complete TriWorldBench Results}
\label{app:triworldbench-complete}

Tables~\ref{tab:triworldbench-complete-1} and~\ref{tab:triworldbench-complete-2} reproduce the complete 36-model leaderboard from the same Sep. 11, 2026 snapshot used in the main text, retaining all 19 original evaluation metrics. VLM Consistency I--III are reported separately. TWB-Score is the official aggregate over all 19 metrics. This appendix includes every leaderboard entry, irrespective of technical-report availability.

\begin{table}[!htbp]
  \centering
  \caption{Complete TriWorldBench leaderboard (part 1 of 2), reporting all 36 models from the Sep. 11, 2026 snapshot. TWB-Score is reproduced from the official leaderboard.}
  \label{tab:triworldbench-complete-1}
  \renewcommand{\baselinestretch}{1}
  \fontsize{8.2}{10}\selectfont
  \setlength{\tabcolsep}{1.7pt}
  \renewcommand{\arraystretch}{1.05}
  \begin{tabularx}{\linewidth}{@{}>{\raggedright\arraybackslash}p{0.19\linewidth}*{9}{>{\centering\arraybackslash}X}!{\vrule width 0.5pt}>{\centering\arraybackslash}p{0.065\linewidth}@{}}
    \toprule[0.7pt]
    \textbf{Model} & {\fontsize{7}{8.4}\selectfont\bfseries\shortstack{Norm.\\PSNR}} & {\fontsize{7}{8.4}\selectfont\bfseries\shortstack{SSIM}} & {\fontsize{7}{8.4}\selectfont\bfseries\shortstack{VLM\\Cons. I}} & {\fontsize{7}{8.4}\selectfont\bfseries\shortstack{VLM\\Cons. II}} & {\fontsize{7}{8.4}\selectfont\bfseries\shortstack{VLM\\Cons. III}} & {\fontsize{7}{8.4}\selectfont\bfseries\shortstack{VQA\\Cons.}} & {\fontsize{7}{8.4}\selectfont\bfseries\shortstack{Instruction\\Following}} & {\fontsize{7}{8.4}\selectfont\bfseries\shortstack{Semantic\\Alignment}} & {\fontsize{7}{8.4}\selectfont\bfseries\shortstack{JEPA\\Similarity}} & \cellcolor{blue!18}{\bfseries\shortstack{TWB-\\Score}} \\
    \midrule[0.35pt]
    \rowcolor{opensourcebg}
    \textbf{CWM} & 72.41 & 83.58 & \underline{85.83} & 84.43 & \underline{95.64} & \underline{70.03} & \underline{76.78} & 90.02 & 87.31 & \cellcolor{blue!18}\textbf{66.04} \\
    dream4act & 70.91 & 84.22 & \textbf{85.99} & \textbf{86.27} & \textbf{96.37} & 66.05 & 74.65 & 90.27 & 87.75 & \cellcolor{blue!7}\underline{65.66} \\
    BWM & \underline{73.43} & \underline{87.47} & 85.11 & 83.94 & 95.31 & 65.94 & 73.84 & 89.93 & \textbf{94.37} & \cellcolor{blue!7}65.54 \\
    WoVR Plus & \textbf{74.62} & \textbf{87.91} & 83.79 & 83.94 & 94.84 & 69.11 & 70.67 & \textbf{90.61} & 87.12 & \cellcolor{blue!7}65.39 \\
    PhyxWM & 70.79 & 85.30 & 84.12 & 76.42 & 90.55 & 59.85 & 73.86 & 90.24 & \underline{91.89} & \cellcolor{blue!7}64.26 \\
    XiaomiAutoWM & 72.04 & 85.72 & 82.94 & 83.17 & 94.53 & 69.27 & 66.40 & 89.94 & 85.71 & \cellcolor{blue!7}63.94 \\
    Fysiverse-Video & 68.90 & 82.68 & 83.39 & \underline{84.54} & 95.05 & 63.32 & 70.18 & \underline{90.34} & 79.59 & \cellcolor{blue!7}63.62 \\
    BetaBWM & 66.87 & 82.54 & 84.77 & 82.21 & 94.77 & 62.26 & 70.87 & 89.34 & 89.08 & \cellcolor{blue!7}63.62 \\
    WMEpivoryn & 67.67 & 82.29 & 80.34 & 79.80 & 92.71 & 59.33 & 67.36 & 89.94 & 79.32 & \cellcolor{blue!7}62.10 \\
    CoWM & 67.67 & 82.25 & 80.36 & 79.92 & 92.51 & 58.34 & 67.41 & 90.19 & 79.06 & \cellcolor{blue!7}62.07 \\
    TrivoryxWM & 67.74 & 82.35 & 77.61 & 77.86 & 91.25 & 53.81 & 66.81 & 89.78 & 73.38 & \cellcolor{blue!7}60.79 \\
    CogVerse & 67.19 & 78.93 & 84.13 & 78.59 & 90.96 & 64.09 & 67.07 & 89.53 & 74.27 & \cellcolor{blue!7}60.49 \\
    Xi-WM-0.4 & 64.08 & 76.50 & 77.57 & 76.00 & 90.16 & 54.70 & 67.83 & 88.58 & 69.12 & \cellcolor{blue!7}59.81 \\
    LibWM & 63.66 & 77.92 & 78.24 & 75.26 & 90.17 & 58.95 & 64.35 & 89.48 & 80.45 & \cellcolor{blue!7}59.29 \\
    ActoVision-X & 67.23 & 81.04 & 82.43 & 80.98 & 92.92 & 60.18 & 65.21 & 89.44 & 67.56 & \cellcolor{blue!7}58.81 \\
    TriCam-Dreamer & 66.69 & 80.69 & 82.71 & 81.37 & 93.18 & 59.52 & 65.36 & 89.34 & 66.96 & \cellcolor{blue!7}58.62 \\
    MirageFlow-V3 & 66.16 & 80.29 & 82.86 & 81.98 & 93.56 & 59.96 & 65.77 & 89.08 & 65.21 & \cellcolor{blue!7}58.43 \\
    DynaWorld & 66.47 & 80.54 & 82.46 & 81.39 & 93.12 & 58.67 & 64.69 & 89.22 & 66.50 & \cellcolor{blue!7}58.42 \\
    RopeWM & 61.93 & 76.49 & 78.93 & 76.14 & 90.86 & 55.48 & 62.41 & 88.52 & 70.08 & \cellcolor{blue!7}57.70 \\
    Ra\_1 & 61.59 & 75.35 & 80.69 & 77.69 & 91.77 & 55.75 & 61.87 & 87.91 & 63.16 & \cellcolor{blue!7}57.25 \\
    SD2-WM & 60.52 & 75.64 & 76.51 & 73.96 & 88.97 & 52.58 & 60.14 & 88.17 & 64.18 & \cellcolor{blue!7}56.38 \\
    LightWM-Alpha & 59.74 & 74.89 & 74.69 & 71.52 & 88.06 & 51.17 & 58.96 & 88.56 & 61.63 & \cellcolor{blue!7}55.62 \\
    Kimo & 59.30 & 75.05 & 72.13 & 69.88 & 87.11 & 49.67 & 58.27 & 88.30 & 65.36 & \cellcolor{blue!7}55.01 \\
    Flash-WM & 57.98 & 75.41 & 73.78 & 72.13 & 88.49 & 61.42 & 56.84 & 88.74 & 45.13 & \cellcolor{blue!7}54.45 \\
    ArmW & 56.57 & 75.63 & 74.16 & 74.71 & 89.68 & \textbf{73.34} & 55.08 & 88.31 & 28.95 & \cellcolor{blue!7}53.88 \\
    WDK-XLAB & 61.28 & 76.88 & 78.96 & 74.81 & 89.94 & 51.46 & 50.62 & 88.06 & 31.84 & \cellcolor{blue!7}53.75 \\
    Ennerverse-AC & 64.61 & 76.97 & 81.42 & 75.25 & 89.70 & 43.67 & 38.80 & 87.73 & 30.11 & \cellcolor{blue!7}53.70 \\
    DreamDojo & 59.78 & 71.58 & 77.14 & 73.26 & 88.19 & 47.82 & 49.72 & 88.28 & 30.73 & \cellcolor{blue!7}51.72 \\
    Dream & 46.60 & 63.46 & 47.16 & 37.82 & 40.46 & 41.04 & 74.01 & 87.79 & 0.79 & \cellcolor{blue!7}49.88 \\
    WMxel & 46.35 & 63.28 & 46.94 & 37.30 & 39.83 & 41.22 & 73.45 & 87.20 & 0.93 & \cellcolor{blue!7}49.56 \\
    LatentGrip & 45.45 & 60.01 & 47.22 & 37.81 & 40.48 & 41.04 & 70.33 & 85.54 & 0.67 & \cellcolor{blue!7}49.24 \\
    Act-Cosmos & 42.76 & 48.60 & 48.56 & 35.69 & 30.45 & 34.66 & \textbf{83.63} & 84.72 & 0.00 & \cellcolor{blue!7}48.93 \\
    Motus & 60.13 & 76.31 & 63.68 & 67.56 & 87.22 & 45.33 & 52.28 & 86.54 & 10.24 & \cellcolor{blue!7}42.35 \\
    Genie Envisioner & 58.74 & 74.44 & 69.40 & 57.13 & 77.93 & 36.72 & 20.20 & 79.30 & 0.02 & \cellcolor{blue!7}40.73 \\
    Ctrl-World & 46.39 & 64.62 & 46.44 & 63.69 & 82.46 & 40.90 & 38.90 & 81.94 & 10.32 & \cellcolor{blue!7}38.98 \\
    KineWorld & 59.31 & 78.35 & 31.94 & 64.90 & 86.23 & 36.53 & 27.50 & 84.72 & 1.77 & \cellcolor{blue!7}38.39 \\
    \bottomrule[0.7pt]
  \end{tabularx}
  \par\vspace{4pt}
  {\fontsize{7.8}{9.4}\selectfont\raggedright
    Bold and underlined scores denote first and second place, respectively, among all 36 models; ties receive the same marking. Rows follow descending TWB-Score. Cons. denotes Consistency. Norm. PSNR denotes Normalized PSNR. All scores are on a 0--100 scale (higher is better).\par}
\end{table}

\clearpage

\begin{table}[!htbp]
  \centering
  \caption{Complete TriWorldBench leaderboard (part 2 of 2), reporting all 36 models from the Sep. 11, 2026 snapshot. TWB-Score is reproduced from the official leaderboard.}
  \label{tab:triworldbench-complete-2}
  \renewcommand{\baselinestretch}{1}
  \fontsize{8.2}{10}\selectfont
  \setlength{\tabcolsep}{1.7pt}
  \renewcommand{\arraystretch}{1.05}
  \begin{tabularx}{\linewidth}{@{}>{\raggedright\arraybackslash}p{0.19\linewidth}*{10}{>{\centering\arraybackslash}X}!{\vrule width 0.5pt}>{\centering\arraybackslash}p{0.065\linewidth}@{}}
    \toprule[0.7pt]
    \textbf{Model} & {\fontsize{7}{8.4}\selectfont\bfseries\shortstack{Interact.\\Quality}} & {\fontsize{7}{8.4}\selectfont\bfseries\shortstack{Persp.}} & {\fontsize{7}{8.4}\selectfont\bfseries\shortstack{State\\Align.}} & {\fontsize{7}{8.4}\selectfont\bfseries\shortstack{Flow\\Score}} & {\fontsize{7}{8.4}\selectfont\bfseries\shortstack{Traj.\\Accuracy}} & {\fontsize{7}{8.4}\selectfont\bfseries\shortstack{Subject\\Cons.}} & {\fontsize{7}{8.4}\selectfont\bfseries\shortstack{Bkgd.\\Cons.}} & {\fontsize{7}{8.4}\selectfont\bfseries\shortstack{Photo.\\Smooth.}} & {\fontsize{7}{8.4}\selectfont\bfseries\shortstack{Image\\Quality}} & {\fontsize{7}{8.4}\selectfont\bfseries\shortstack{Aesthetic\\Quality}} & \cellcolor{blue!18}{\bfseries\shortstack{TWB-\\Score}} \\
    \midrule[0.35pt]
    \rowcolor{opensourcebg}
    \textbf{CWM} & 31.26 & \textbf{91.16} & 61.01 & 22.99 & 42.98 & \underline{84.55} & 75.54 & 30.28 & \textbf{43.24} & 25.76 & \cellcolor{blue!18}\textbf{66.04} \\
    dream4act & 36.08 & 86.51 & 60.73 & 24.10 & 40.16 & 83.27 & 76.01 & 35.36 & 37.60 & 25.24 & \cellcolor{blue!7}\underline{65.66} \\
    BWM & 33.59 & 87.20 & 61.42 & 20.48 & 41.97 & 84.20 & 70.29 & 33.95 & 38.71 & 24.13 & \cellcolor{blue!7}65.54 \\
    WoVR Plus & 28.76 & 85.49 & 60.91 & \underline{24.85} & \underline{45.27} & 84.22 & 75.19 & 32.90 & 36.22 & 26.06 & \cellcolor{blue!7}65.39 \\
    PhyxWM & 28.41 & 86.19 & 60.80 & 20.34 & \textbf{45.28} & 84.30 & 75.12 & 35.32 & 38.13 & 23.99 & \cellcolor{blue!7}64.26 \\
    XiaomiAutoWM & 29.66 & 78.28 & 60.84 & 23.18 & 42.85 & 83.61 & 75.23 & 33.34 & 33.63 & 24.50 & \cellcolor{blue!7}63.94 \\
    Fysiverse-Video & 32.19 & 83.86 & 60.82 & 22.16 & 37.40 & 83.65 & 75.35 & 35.63 & 36.17 & 23.62 & \cellcolor{blue!7}63.62 \\
    BetaBWM & 33.14 & 83.77 & 60.14 & 21.00 & 34.67 & 83.38 & 75.22 & 35.40 & 36.42 & 22.87 & \cellcolor{blue!7}63.62 \\
    WMEpivoryn & 27.21 & 84.90 & 60.57 & 20.21 & 35.90 & 84.06 & 73.58 & 36.22 & 34.53 & 23.98 & \cellcolor{blue!7}62.10 \\
    CoWM & 27.46 & 85.21 & 60.68 & 19.59 & 35.98 & 84.21 & 73.03 & 36.61 & 34.85 & 23.96 & \cellcolor{blue!7}62.07 \\
    TrivoryxWM & 26.53 & 83.80 & 60.81 & 18.29 & 34.59 & 83.77 & 72.94 & 37.73 & 33.15 & 22.89 & \cellcolor{blue!7}60.79 \\
    CogVerse & 28.59 & 76.80 & 60.04 & 17.56 & 29.53 & 83.64 & 72.68 & 36.01 & 27.82 & 21.86 & \cellcolor{blue!7}60.49 \\
    Xi-WM-0.4 & 27.58 & 81.87 & 61.05 & 23.71 & 30.10 & 82.60 & 75.37 & 28.88 & 38.45 & 22.21 & \cellcolor{blue!7}59.81 \\
    LibWM & 27.00 & 72.82 & 60.58 & 17.38 & 30.06 & 83.61 & 74.45 & 35.59 & 25.68 & 20.79 & \cellcolor{blue!7}59.29 \\
    ActoVision-X & 30.38 & 77.73 & 61.75 & 17.68 & 28.46 & 82.89 & 58.37 & 23.33 & 28.20 & 21.62 & \cellcolor{blue!7}58.81 \\
    TriCam-Dreamer & 30.61 & 76.96 & 61.66 & 17.77 & 26.73 & 82.84 & 58.76 & 22.99 & 28.14 & 21.44 & \cellcolor{blue!7}58.62 \\
    MirageFlow-V3 & 31.00 & 76.82 & 61.81 & 17.54 & 26.34 & 82.83 & 56.86 & 23.09 & 27.64 & 21.33 & \cellcolor{blue!7}58.43 \\
    DynaWorld & 31.36 & 76.90 & 61.81 & 17.94 & 26.28 & 82.84 & 57.94 & 22.89 & 27.61 & 21.31 & \cellcolor{blue!7}58.42 \\
    RopeWM & 27.26 & 71.35 & 60.32 & 16.07 & 24.16 & 82.94 & 73.24 & 34.66 & 25.51 & 19.95 & \cellcolor{blue!7}57.70 \\
    Ra\_1 & 26.23 & 73.22 & 60.45 & 15.88 & 19.76 & 83.89 & 71.46 & 35.73 & 25.38 & 20.06 & \cellcolor{blue!7}57.25 \\
    SD2-WM & 25.49 & 70.58 & 60.71 & 15.19 & 21.65 & 83.32 & 72.76 & 36.47 & 24.93 & 19.45 & \cellcolor{blue!7}56.38 \\
    LightWM-Alpha & 24.98 & 68.27 & 60.37 & 13.94 & 20.18 & 82.15 & 74.62 & 34.38 & 29.59 & 19.08 & \cellcolor{blue!7}55.62 \\
    Kimo & 24.60 & 66.87 & 60.46 & 14.52 & 20.79 & 83.83 & 71.33 & 36.13 & 22.76 & 18.79 & \cellcolor{blue!7}55.01 \\
    Flash-WM & 26.91 & 64.92 & 61.18 & 11.26 & 18.15 & 80.16 & 80.67 & 28.79 & 24.03 & 18.56 & \cellcolor{blue!7}54.45 \\
    ArmW & 27.69 & 63.66 & 60.84 & 6.50 & 16.08 & 77.35 & \underline{89.23} & 19.27 & 27.88 & 18.77 & \cellcolor{blue!7}53.88 \\
    WDK-XLAB & 26.45 & 68.43 & \underline{62.41} & 21.86 & 16.89 & 75.14 & 82.97 & 24.79 & 20.12 & 18.34 & \cellcolor{blue!7}53.75 \\
    Ennerverse-AC & 26.17 & 72.17 & \textbf{63.03} & \textbf{35.39} & 16.38 & 74.09 & 81.35 & 23.20 & 22.00 & 18.23 & \cellcolor{blue!7}53.70 \\
    DreamDojo & 28.28 & 59.39 & 60.76 & 6.52 & 16.61 & 77.37 & \textbf{89.29} & 15.86 & 24.45 & 17.58 & \cellcolor{blue!7}51.72 \\
    Dream & \underline{69.14} & 88.63 & 58.92 & 7.53 & 11.15 & 80.52 & 72.31 & 56.60 & 37.85 & 25.98 & \cellcolor{blue!7}49.88 \\
    WMxel & 67.73 & 88.96 & 59.25 & 7.76 & 11.05 & 80.46 & 71.10 & 55.58 & 37.23 & \underline{26.09} & \cellcolor{blue!7}49.56 \\
    LatentGrip & 66.16 & 88.85 & 59.24 & 5.06 & 11.09 & 81.64 & 68.70 & \textbf{60.37} & 40.17 & 25.66 & \cellcolor{blue!7}49.24 \\
    Act-Cosmos & \textbf{71.34} & \underline{90.43} & 58.09 & 6.88 & 13.28 & 74.25 & 76.87 & \underline{60.21} & \underline{40.20} & \textbf{29.04} & \cellcolor{blue!7}48.93 \\
    Motus & 25.00 & 44.20 & 59.31 & 0.80 & 13.77 & 77.37 & 1.00 & 1.31 & 16.08 & 16.44 & \cellcolor{blue!7}42.35 \\
    Genie Envisioner & 20.10 & 87.91 & 59.28 & 0.40 & 0.84 & \textbf{91.40} & 1.07 & 4.08 & 19.36 & 15.57 & \cellcolor{blue!7}40.73 \\
    Ctrl-World & 24.19 & 45.15 & 60.47 & 21.29 & 6.09 & 77.05 & 49.02 & 12.44 & 18.37 & 14.92 & \cellcolor{blue!7}38.98 \\
    KineWorld & 22.00 & 52.55 & 59.19 & 0.25 & 4.08 & 84.06 & 0.73 & 2.21 & 17.60 & 15.42 & \cellcolor{blue!7}38.39 \\
    \bottomrule[0.7pt]
  \end{tabularx}
  \par\vspace{4pt}
  {\fontsize{7.8}{9.4}\selectfont\raggedright
    Bold and underlined scores denote first and second place, respectively, among all 36 models; ties receive the same marking. Rows follow descending TWB-Score. Cons. denotes Consistency. Align. = Alignment; Interact. = Interaction; Persp. = Perspective; Traj. = Trajectory; Bkgd. = Background; Photo. Smooth. = Photometric Smoothness. All scores are on a 0--100 scale (higher is better).\par}
\end{table}
\clearpage
\section{Detailed PAI-Bench-G Robot Domain (RO) Results}
\label{app:paibench-g-details}

\begin{table}[!htbp]
  \centering
  \caption{Locally evaluated RO subcategory scores on PAI-Bench-G, grouped by the original question labels. CausalWM (CWM) and Cosmos3-Super are evaluated using the official evaluation protocol with Qwen3-VL-235B-A22B-Instruct~\cite{qwen3vl} as judge. Subcategory scores report mean per-video Visual Question Answering (VQA) accuracy (\%) over questions of the corresponding type, averaged across samples containing that type and five seeds. Overall RO uses all question types. Scores are reported on a 0--100 scale.}
  \label{tab:paibench-g-breakdown}
  \renewcommand{\baselinestretch}{1}
  \fontsize{8.8}{11}\selectfont
  \setlength{\tabcolsep}{6pt}
  \renewcommand{\arraystretch}{1.3}
  \begin{minipage}{0.86\linewidth}
  {\fontsize{9.5}{11}\selectfont\bfseries Breakdown by original question type\par}
  \vspace{4pt}
  \begin{tabularx}{\linewidth}{@{\hspace{6pt}}
    >{\raggedright\arraybackslash}Xrr
    >{\centering\arraybackslash}p{0.20\linewidth}!{\vrule width 0.5pt}
    >{\centering\arraybackslash}p{0.20\linewidth}@{\hspace{6pt}}}
    \toprule[0.7pt]
    {\fontsize{9.5}{11}\selectfont\bfseries Question type}
    & {\fontsize{9.5}{11}\selectfont\bfseries Samples}
    & {\fontsize{9.5}{11}\selectfont\bfseries Questions}
    
    & {\fontsize{9.5}{11}\selectfont\bfseries Cosmos3-Super}
    & \cellcolor{opensourcebg}{\fontsize{9.5}{11}\selectfont\bfseries CWM (Ours)} \\
    \midrule[0.35pt]
    \rowcolor{gray!12}
    \multicolumn{5}{@{\hspace{6pt}}l@{\hspace{6pt}}}{\fontsize{9.5}{11}\selectfont\bfseries Physics} \\
    Attributes & 102 & 122 & {\fontsize{9}{11}\selectfont \textbf{95.1}} & \cellcolor{opensourcebg}{\fontsize{9}{11}\selectfont \underline{94.6}} \\
    Object Permanence & 73 & 79 & {\fontsize{9}{11}\selectfont \underline{85.6}} & \cellcolor{opensourcebg}{\fontsize{9}{11}\selectfont \textbf{86.8}} \\
    States & 87 & 107 & {\fontsize{9}{11}\selectfont \underline{90.0}} & \cellcolor{opensourcebg}{\fontsize{9}{11}\selectfont \textbf{92.1}} \\
    \midrule[0.35pt]
    \rowcolor{gray!12}
    \multicolumn{5}{@{\hspace{6pt}}l@{\hspace{6pt}}}{\fontsize{9.5}{11}\selectfont\bfseries Space} \\
    Geometry & 58 & 74 & {\fontsize{9}{11}\selectfont \textbf{94.7}} & \cellcolor{opensourcebg}{\fontsize{9}{11}\selectfont \underline{91.2}} \\
    Interaction & 99 & 130 & {\fontsize{9}{11}\selectfont \textbf{91.7}} & \cellcolor{opensourcebg}{\fontsize{9}{11}\selectfont \underline{91.3}} \\
    Relationship & 99 & 126 & {\fontsize{9}{11}\selectfont \underline{90.6}} & \cellcolor{opensourcebg}{\fontsize{9}{11}\selectfont \textbf{94.8}} \\
    \midrule[0.35pt]
    \rowcolor{gray!12}
    \multicolumn{5}{@{\hspace{6pt}}l@{\hspace{6pt}}}{\fontsize{9.5}{11}\selectfont\bfseries Time} \\
    Action & 107 & 130 & {\fontsize{9}{11}\selectfont \textbf{89.3}} & \cellcolor{opensourcebg}{\fontsize{9}{11}\selectfont \underline{86.6}} \\
    Camera & 62 & 65 & {\fontsize{9}{11}\selectfont \underline{92.7}} & \cellcolor{opensourcebg}{\fontsize{9}{11}\selectfont \textbf{93.9}} \\
    Order & 67 & 80 & {\fontsize{9}{11}\selectfont \textbf{86.6}} & \cellcolor{opensourcebg}{\fontsize{9}{11}\selectfont \underline{83.9}} \\
    \midrule[0.35pt]
    \rowcolor{blue!7}
    \textbf{Overall RO} & 174 & 913 & {\fontsize{9}{11}\selectfont \underline{89.7}} & \cellcolor{blue!18}{\fontsize{9}{11}\selectfont \textbf{89.9}} \\
    \bottomrule[0.7pt]
  \end{tabularx}

  \end{minipage}
\end{table}

\end{document}